\documentclass{article}

\usepackage[main, final]{neurips_2026}
\usepackage{wrapfig}
\usepackage{makecell}

\usepackage[utf8]{inputenc} 
\usepackage[T1]{fontenc}    
\usepackage{tcolorbox}
\usepackage[normalem]{ulem}

\definecolor{darkblue}{rgb}{0,0.25,0.75}
\definecolor{mqarcontent}{HTML}{eb5e28}
\definecolor{mqardelim}{HTML}{403d39}
\definecolor{mqarnoise}{HTML}{ccc5b9}

\usepackage{xcolor}
\usepackage{listings}

\lstdefinestyle{pythonstyle}{
    language=Python,
    basicstyle=\ttfamily\small,
    keywordstyle=\color{blue},
    commentstyle=\color{gray},
    stringstyle=\color{teal},
    numbers=left,
    numberstyle=\tiny\color{gray},
    stepnumber=1,
    numbersep=8pt,
    showstringspaces=false,
    breaklines=true,
    breakatwhitespace=true,
    frame=single,
    captionpos=b,
    tabsize=4
}

\usepackage{hyperref}
\hypersetup{
  colorlinks=true,
  linkcolor=darkblue,
  citecolor=darkblue,
  urlcolor=darkblue
}

\usepackage{url}            
\usepackage{booktabs}       
\usepackage{amsfonts}       
\usepackage{nicefrac}       
\usepackage{microtype}      
\usepackage{xcolor}         
\usepackage{graphicx}
\usepackage{multirow}
\usepackage{array}
\usepackage{subcaption} 
\usepackage{amsmath}
\title{Dynamic Short Convolutions Improve Transformers}

\author{
\textbf{Oliver Sieberling}$^{1}$ \quad \textbf{Bharat Runwal}$^{2}$ \quad \textbf{Rameswar Panda}$^{2}$ \quad \textbf{Yoon Kim}$^{1}$ \vspace{2mm} \\ 
$^{1}$Massachusetts Institute of Technology \quad
$^{2}$MIT-IBM Watson AI Lab \vspace{2mm} \\
\texttt{osieberl@mit.edu}
}

\begin{document}

\maketitle

\begin{abstract}
  Transformers have become the dominant architecture for large language models, largely due to the scalability and flexibility of attention, feed-forward layers, residual connections, and normalization. This paper introduces dynamic short convolutions as an additional neural network primitive for improving Transformers. Unlike static short convolutions, dynamic convolutions use input-dependent filters, which preserves the locality bias of convolution while increasing expressivity. Motivating experiments show that applying dynamic short convolutions to key, query, and value representations improves performance on challenging associative recall tasks compared with static convolutional variants. Across language-modeling experiments ranging from 150M to 2B parameters, dynamic convolutions consistently outperform standard Transformers and Transformers augmented with static short convolutions. Fitting scaling laws indicates a 1.33$\times$ compute advantage over compute-matched Transformers when dynamic convolutions are applied to the key, query, value vectors, and a 1.60$\times$ advantage when adding dynamic convolutions after every linear layer. Dynamic convolutions also offer improvements on linear RNNs (Mamba-2/Gated DeltaNet) and mixture-of-experts architectures. We make these gains practical with custom Triton kernels\footnote{The Triton kernels are available at \url{https://github.com/OliverSieberling/dynamic-conv1d}.} that enable efficient training with a manageable end-to-end slowdown.  These results suggest that dynamic short convolutions are a scalable, hardware-efficient, and expressive primitive for advancing Transformer-based language models.
\end{abstract}

\section{Introduction}
Individual neural network layers form the primitive building blocks of deep learning architectures. Core primitives that have become  mainstays  include multilayer perceptrons \citep{rosenblatt1958perceptron,rumelhart1986learning}, convolutions \citep{fukushima1980neocognitron,lecun1998gradient}, recurrent layers  \citep{elman1990finding,hochreiter1997lstm,cho2014learning} and attention \citep{bahdanau2014nmtattention}. Residual connections \citep{he2016deep} and normalization techniques \citep{ioffe2015batchnorm,ba2016layernorm} are also crucial for  practical training of deep architectures built out of such layers. 

The Transformer architecture \citep{vaswani2017attention} exemplifies how such  primitives can be composed into a flexible and scalable model that is  effective across domains. Transformers are built from repeatedly interleaving attention and feed-forward blocks with residual connections and layer normalization. Major refinements to these components since  inception include gated and mixture-of-experts feed-forward layers \citep{shazeer2020glu,fedus2022switch},  placement/type of normalization layers \citep{zhang2019rmsnorm,xiong2020layernorm},   key-value sharing techniques \citep{shazeer2019mqa,ainslie2023gqa}, and relative positional encodings  \citep{su2021roformer}. These architectural advances, combined with improved optimization techniques and higher quality training data, have significantly pushed the performance-efficiency frontier---indeed, modern sub-10B-parameter LLMs routinely outperform older  100B+ parameter LLMs based on older Transformer variants.

This paper proposes \emph{dynamic  convolutions} as an additional primitive for improving the Transformer. Convolution layers, which apply a shared local filter across sequence positions to mix neighboring token representations, have long been used as the  ``sequence mixing'' component in deep models for natural language processing, from early seminal work on word-level tagging \citep{collobert2008unified,collobert2011nlp}, to sentence-level classification \citep{kalchbrenner2014convolutional,kim2014convolutional},  sequence-to-sequence learning \citep{kalchbrenner2016neural,gehring2017convolutional}, and language modeling \citep{dauphin2017language}. However, they largely fell out of favor as a primary sequence-mixing mechanism following the introduction of Transformers. In the post-Transformer era, some works have instead found that incorporating  lightweight depthwise-separable convolution layers that apply independent local filters within each channel (also called \emph{short convolutions}) into Transformers can improve performance in some settings \citep{so2021primer,allen2025physics}. To the best of our knowledge, however, such layers  are generally not part of frontier open-weight LLMs.\footnote{Short convolution layers are, however, standard in recent linear RNNs such as Mamba \citep{Gu2023MambaLS,dao2024transformers} and DeltaNet \citep{schlag_linear_2021,yang2024parallelizing,yang2024gated}, though even more  recent linear RNNs such as Mamba-3 \citep{lahoti2026mamba} and Raven \citep{afzalbick2026raven} eschew the use of short convolutions.}

Dynamic (short) convolutions generalize short convolutions by allowing the convolutional filter at each time step to depend on the input, for example by parameterizing it as a learned linear transformation of the current hidden state \citep{jia2016dynamic, wu2019pay, chen2020dynamicconv, jiang2020convbert, gu2026jet}. This input-dependent parameterization preserves the locality bias of convolutions while  increasing their expressivity. For a layer to be practically useful, however, increased expressivity is not enough---it must also be \emph{scalable}. Scalability in the context of modern LLMs means several things. For one, the layer should continue to provide improvements as the model and training data are scaled up. Two, insofar as new layers typically introduce more compute/parameters, the new layer should increase the overall rate at which an architecture can trade off resources for performance, i.e., it should outperform the existing architecture when compute-/parameter-matched. Finally, the layer should be hardware-efficient, i.e., efficiently trainable on modern accelerators such as GPUs and TPUs.

We show that dynamic short convolutions satisfy the above desiderata. Across experiments spanning models with 150M-2B  parameters, dynamic convolutions consistently improve upon Transformers with and without short convolutions. Fitting scaling law curves to the results suggests that dynamic convolutions offer a 1.33$\times$ compute advantage compared to ordinary  Transformers when the dynamic convolutions are applied to the QKV layers, and a 1.60$\times$ advantage when they are applied to all linear layers. For wall-clock efficiency, we develop a Triton kernel that results in competitive performance with a well-optimized static short convolution kernel. Combined with an efficient input-dependent filter parameterization, the end-to-end training throughput slowdown is manageable: roughly 8\% slowdown for the QKV variant and 22\% slowdown for the all-linear variant at the 2B scale. These results collectively position dynamic convolutions as an additional primitive to be considered for improving the Transformer.

\section{Dynamic Short Convolutions for Transformers}

\subsection{Parameterization}
A short convolution is a depthwise separable convolution \citep{chollet2017xception,howard2017mobilenets} with kernel width $W$ (typically $W \in \{3, 4, 5\}$ in language applications), applied along the time dimension. More specifically, a static short convolution computes:
\begin{equation}
    y_t := \sum_{k=0}^{W-1} w_k \odot x_{t-k},
\end{equation}
where $x_t \in \mathbb{R}^{D}$ is a sequence of activations, $w \in \mathbb{R}^{W\times D}$ is the convolution filter, and $\odot$ denotes an elementwise product. Note that here the convolution weights $w$ are fixed across the time axis. 

Dynamic short convolutions generalize static short convolutions by making the convolution kernel input-dependent. At each position $t$, a weight generator (e.g., a linear projection) produces the dynamic convolution weights $w^{(t)} \in \mathbb{R}^{W \times D}$, and the convolution is performed with this time-varying filter:
\begin{equation}
    y_t := \sum_{k=0}^{W-1} w^{(t)}_k \odot x_{t-k}.
\end{equation}
Each token thus selects its own filter to retrieve information from the local context. In this respect, dynamic convolutions are reminiscent of attention, but instead of deriving the attention weights from query-key similarity, dynamic convolutions generate them directly from the querying position \citep{wu2019pay}. While this mechanism does not reference the content being retrieved, it carries a strong inductive bias toward retrieving by relative position within the filter window.

While dynamic convolutions are expressive, na\"ively producing the dynamic convolution weights would require a $D \rightarrow W \cdot D$ linear projection, which would roughly double the parameter count of the underlying model (assuming $W=4$). We therefore consider more parameter-efficient parameterizations. In our first approach, we   factorize the projection through a low-rank transformation of rank $R$. In our second approach, we split the dimensions into different ``heads'', using the  transformation $D \rightarrow W \cdot (D/H)$  and broadcasting each weight across a head of size $H$. While we generally found the low-rank parameterization to perform better, the head-wise variant simplifies the design of an efficient GPU kernel. We also have a bias in the above transformations, and hence the filters are affine transformations of the input.

For placement, for our main experiments we place the dynamic short convolutions on the queries, keys, and values before RoPE, with kernel width $W=4$. We apply each with a residual, i.e., $X = X + \mathrm{dynamicShortConv}(X)$ for $X \in \{Q, K,V\}$. The projection that generates the dynamic convolution weights takes the post-attention-norm activations as input.\footnote{We found this to perform slightly better than taking $Q$, $K$, $V$ themselves as input, and it allows the projection to be fused with the $\texttt{qkv\_projection}$.} We also experiment with placing dynamic convolutions after all linear layers of a Transformer.

\subsection{Efficient Training}
Dynamic short convolutions have low arithmetic intensity and are therefore bound by memory accesses. Na\"ive PyTorch implementations repeatedly move intermediate tensors to and from HBM, making dynamic convolutions slow in practice.
We address this with a custom Triton \citep{tillet2019triton} kernel that takes the activations and dynamic convolution weights as input, performs the full convolution on-chip, and writes only the final result back to HBM. Each input is read once and each output is written once, so performance is limited primarily by HBM bandwidth.

Since the dynamic-weight tensor of shape $B \times T \times D \times W$ is $W$ times larger than the $B \times T \times D$ activation tensor, it dominates HBM traffic. Therefore, reducing the size of the dynamic convolution weights translates directly into lower latency. Our head-wise dynamic convolution, which shares a single weight filter across $H$ consecutive channels (head-wise tying),  reduces the dynamic-weight tensor to $B \times T \times (D/H) \times W$. When $H \gg W$, its IO cost becomes negligible relative to the activations.

For the low-rank prediction of the dynamic convolution weights, we develop a separate Triton kernel that fuses the second projection of a low-rank factorization directly into the convolution kernel. Rather than reading the materialized $B \times T \times D \times W$ dynamic convolution weights, the kernel reads the $B \times T \times R$ low-rank inputs $z$ and the $R \times (D \cdot W)$ second projection $U$, generates the dynamic weights $zU$ on-chip and immediately applies the convolution. The dynamic weights are never written to HBM, which makes the low-rank kernel significantly faster than the head-size-$1$ kernel.

\begin{figure}[t]

    \centering
    \vspace{2mm}
    \includegraphics[width=1.02\textwidth]{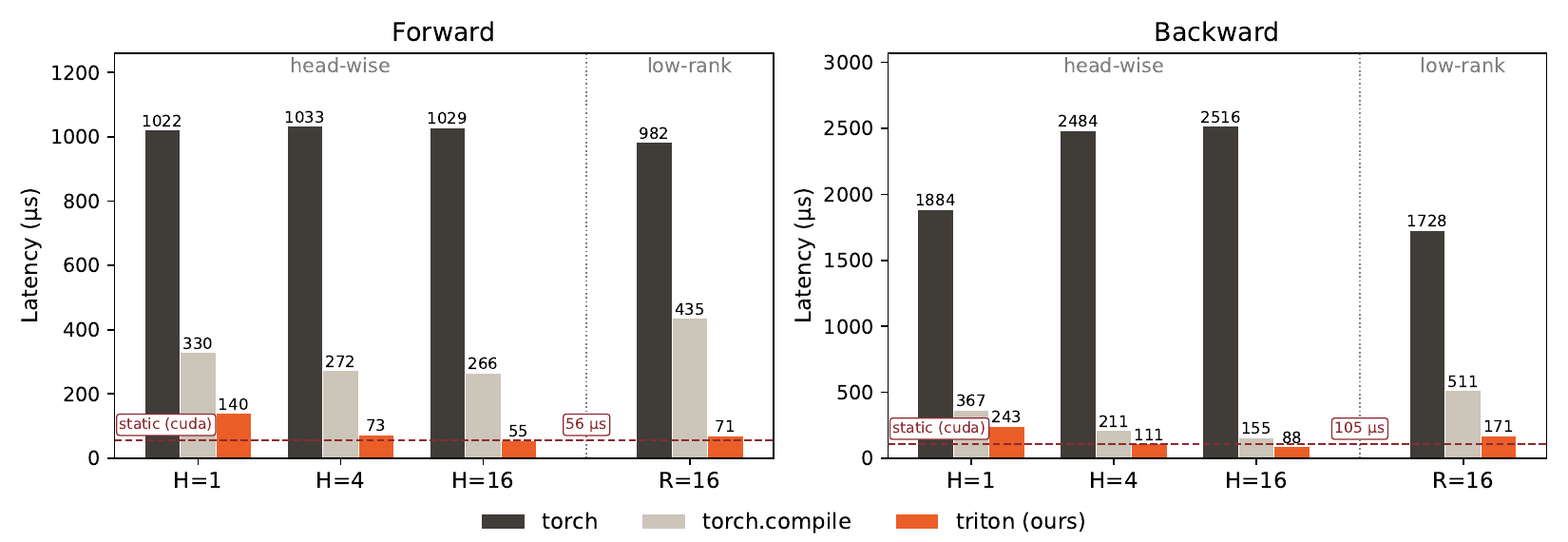}
    \caption{Latency of dynamic short-convolution kernels on an H100 HBM3 80GB GPU ($B=4$, $T=4096$, $D=2048$, $W=4$, BF16). Triton kernels (orange) vs. the best PyTorch eager (dark grey) and \texttt{torch.compile} (light grey) baselines, where each baseline is the fastest of five different implementations. The dashed line is the CUDA-optimized \texttt{causal\_conv1d} kernel for static short-convolutions of the same width from \url{https://github.com/Dao-AILab/causal-conv1d}.}
    \label{fig:kernel}
    \vspace{2mm}
\end{figure}

Figure~\ref{fig:kernel} compares our custom Triton kernels with PyTorch eager and \texttt{torch.compile} baselines \citep{paszke2019pytorch, ansel2024pytorch2}, each selected as the fastest of five mathematically equivalent implementations. We provide a detailed description of the benchmarking setup and the tested baselines in Appendix~\ref{app:kernel_baselines}.
Across all four configurations, our Triton kernels are $1.8$--$3.9\times$ faster than the best \texttt{torch.compile} baseline on the combined forward and backward pass. As expected, the latency decreases as the head-size increases. At $H=16$, the kernel is even faster than the CUDA-optimized implementation for static convolutions, which we believe is due to the simpler reduction of the convolution-weight gradient. The head-wise kernels sustain $2.6$--$3.0$~TB/s of HBM traffic, compared with a theoretical peak of $3.35$~TB/s. Our low-rank kernel has lower latency than the head-size-$1$ kernel despite fusing an additional linear projection, which demonstrates the benefit of avoiding materialization of the dynamic convolution weights. Nevertheless, the low-rank kernel remains less optimized than the head-size-$16$ variant, and an optimized CUDA implementation could further reduce its latency. Overall, our dynamic short convolution kernel is only moderately slower (and for $H \ge 16$ slightly faster) than the CUDA-optimized short convolution kernel\footnote{\url{https://github.com/Dao-AILab/causal-conv1d}}, which to the best of our knowledge is among the state-of-the-art kernels for static short convolutions.

\section{Empirical Study}
We experimentally validate augmenting Transformers with dynamic short convolutions in both synthetic benchmarks and real-world language modeling settings. 

\subsection{Synthetic Benchmarks}

One motivation for dynamic convolutions in language applications is that  language phenomena often require local context-dependent composition to extract meaning from surface form text. For example, consider the phrases 
``\texttt{the old can opener}'' and ``\texttt{the old can swim}''. The first phrase is a noun phrase with the syntactic structure \texttt{[the [old [can opener]]]} while the second is a verb phrase with the structure \texttt{[[the old] [can swim]]}. Even though the prefix of the two phrases is identical, the local composition function over the 4-word window is a function of the last word ``\texttt{opener}'' vs. ``\texttt{swim}''. Successive attention layers can in principle use positional information to compose local context in a context-dependent, dynamic way. However, this is costly and lacks an inductive bias towards locality. Static short convolutions, on the other hand, have a locality bias but do not explicitly model dynamic compositions. Dynamic convolutions are ideally suited for modeling such phenomena. 

We  study such phenomena in a synthetic setting by considering a modified version of the multi-query associative recall \citep[MQAR;][]{arora2023zoology} task. Standard MQAR provides a sequence of $(\texttt{key}, \texttt{value})$ pairs, each appearing twice, and supervises the model to predict the value corresponding to a key at the second appearance of each pair. We modify this task by letting each key consist of a variable number of tokens $L_k \in \{1,2,3\}$, followed by a delimiter token that encodes $L_k$, and a single value token. A short illustrative example is given by:

\begin{center}
\resizebox{0.85\linewidth}{!}{$\displaystyle
\overbrace{\textcolor{mqarcontent}{\texttt{a b c}}\,\textcolor{mqardelim}{\texttt{<3>}}\,\textcolor{mqarcontent}{\texttt{x}}}^{(k_1, v_1)}\ 
\textcolor{mqarnoise}{\texttt{x}}\ 
\overbrace{\textcolor{mqarcontent}{\texttt{b a}}\,\textcolor{mqardelim}{\texttt{<2>}}\,\textcolor{mqarcontent}{\texttt{y}}}^{(k_2, v_2)}\ 
\textcolor{mqarnoise}{\texttt{a}}\ 
\overbrace{\textcolor{mqarcontent}{\texttt{b c}}\,\textcolor{mqardelim}{\texttt{<2>}}\,\textcolor{mqarcontent}{\texttt{z}}}^{(k_3, v_3)}\ 
\overbrace{\textcolor{mqarcontent}{\texttt{b a}}\,\textcolor{mqardelim}{\texttt{<2>}}\,\textcolor{mqarcontent}{\mbox{\uline{\texttt{y}}}}}^{(k_2, v_2)}\ 
\textcolor{mqarnoise}{\texttt{a}}\ 
\overbrace{\textcolor{mqarcontent}{\texttt{b c}}\,\textcolor{mqardelim}{\texttt{<2>}}\,\textcolor{mqarcontent}{\mbox{\uline{\texttt{z}}}}}^{(k_3, v_3)}\ 
\textcolor{mqarnoise}{\texttt{x a}}\ 
\overbrace{\textcolor{mqarcontent}{\texttt{a b c}}\,\textcolor{mqardelim}{\texttt{<3>}}\,\textcolor{mqarcontent}{\mbox{\uline{\texttt{x}}}}}^{(k_1, v_1)}
$}
\end{center}
Concretely, in this example, one key (\texttt{bc}) is a suffix of another key (\texttt{abc}), and therefore a successful retrieval requires a dynamic filter.
Here we have three key-value pairs: $(\texttt{abc}, \texttt{x})$, $(\texttt{ba}, \texttt{y})$, $(\texttt{bc}, \texttt{z})$. The second occurrence of each value (underlined) is the supervision target. In between key-value pairs, there can be random filler tokens (grey). The difficulty of this task is that depending on the delimiter token, a different number of preceding tokens must be aggregated to form the key. \texttt{<3>} indicates that the key is the previous three tokens, \texttt{<2>} the previous two. Because the keys share tokens and have different lengths, no static filter can separate them. 

\begin{figure}[t]

\centering

\begin{minipage}[t]{0.49\textwidth}
    \centering
    \vspace{0mm}
    \includegraphics[width=\linewidth]{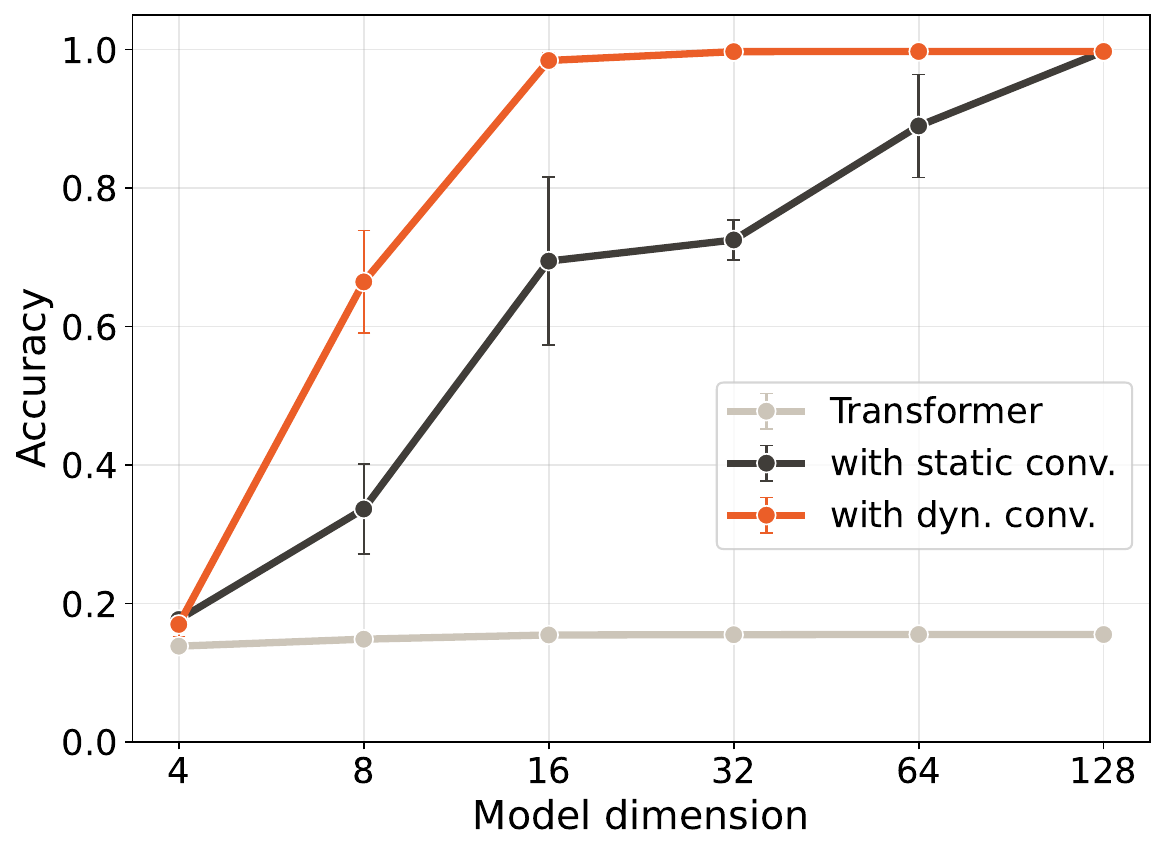}
\end{minipage}
\hfill
\begin{minipage}[t]{0.49\textwidth}
    \centering
    \vspace{3mm}
\resizebox{\linewidth}{!}{%
\begin{tabular}{l ccc}
    \toprule
    \textbf{Task}
        & \textbf{Transformer}
        & \textbf{w/ static}
        & \textbf{w/ dyn.} \\
        &
        & \textbf{conv.}
        & \textbf{conv.} \\
    \midrule
    Compress             & 0.375          & 0.417          & \textbf{0.424} \\
    Fuzzy Recall         & 0.298          & 0.505          & \textbf{0.726} \\
    In-Context Recall    & 0.942          & \textbf{1.000} & \textbf{1.000} \\
    Memorize             & 0.791          & \textbf{0.856} & 0.795          \\
    Noisy Recall         & 0.917          & \textbf{1.000} & \textbf{1.000} \\
    Selective Copy       & 0.930          & 0.983          & \textbf{0.988} \\
    \midrule
    \textbf{Average}     & 0.709          & 0.793          & \textbf{0.822} \\
    \bottomrule
\end{tabular}}
\label{tab:mad}

\end{minipage}

\caption{Left: Performance (median over 5 seeds) on the synthetic variable-key MQAR task. The error bars depict the minimum and maximum values. Right: Performance on the MAD benchmark.}
\label{fig:both_synthetic}
    \vspace{2mm}
\end{figure}

We train Transformers on this task with a single layer and head, varying the model dimension. We compare a vanilla Transformer, the same Transformer with static convolutions on $Q$, $K$, $V$, and our low-rank ($R=16$) dynamic convolution variant. The convolution widths are all set to $W=4$, which is just enough to cover the entire key. We train on $100{,}000$ examples, and report the median accuracy over five seeds. As shown in Figure~\ref{fig:both_synthetic}, Transformers augmented with dynamic short convolutions outperform Transformers with and without static short convolutions for a given model size, highlighting the benefits of input-dependent and local composition functions.

We next test dynamic convolutions on the mechanistic architecture design benchmark \citep[MAD;][]{poli2024mechanistic}, a diagnostic benchmark designed to test the capabilities of different architectures. The results are shown in Figure~\ref{fig:both_synthetic}, where we observe dynamic convolutions (also with $R=16$) to perform well. The improvements are particularly pronounced on the Fuzzy Recall task, where the model must perform in-context recall in a setting where the keys and values consist of a variable number of tokens.

\vspace{-0.5mm}
\subsection{Language Modeling}

We test whether augmenting Transformers (including Mixture of Experts variants) with dynamic short convolutions improves real-world language modeling. We then transfer the same recipe to two strong linear attention variants (Gated DeltaNet \citep{schlag_linear_2021,yang2024parallelizing, yang2024gated} and Mamba-2 \citep{dao2024transformers}).

        \vspace{-2mm}
\paragraph{Experimental Setup.}

We train all models in the \texttt{lm-engine} codebase \citep{mishra2024lmengine} on the Nemotron-CC corpus \citep{su-etal-2025-nemotron} tokenized with the Granite-4 BPE tokenizer (vocabulary 100,352). All runs use sequence length $4096$, RMSNorm \citep{zhang2019rmsnorm}, SwiGLU MLPs \citep{shazeer2020glu}, RoPE \citep{su2021roformer}, and a Llama-style pre-norm block.

For optimization we use AdamW \citep{loshchilov2019decoupledweightdecayregularization} with peak learning rate $3 \times 10^{-4}$, weight decay $0.1$, and learning rate scheduling with $10\%$ warmup and cosine decay to zero. We train dense models at \{150M, 300M, 600M, 1B, 2B\} parameters, with a token-to-parameter ratio of approximately $50$, i.e., $2.5\times$ the compute-optimal recipe recommended by \citet{hoffmann2022training}. Additionally, we train a $7\mathrm{B}$ ($1\mathrm{B}$ active) parameter Mixture of Experts (MoE) model \citep{shazeer2017outrageously} on $100\mathrm{B}$ tokens. A more detailed description of the hyperparameter setting can be found in Appendix~\ref{app:detailed_hyperparams}.

\paragraph{Scaling laws.} We first study the scaling trends of Transformers with and without dynamic convolutions, where for the dynamic convolution we use the low-rank version with ranks $R=\{20,26,32,42,52\}$ for the $\{150M,300M,600M,1B,2B\}$-parameter models.\footnote{These ranks are selected so that the low-rank version roughly matches the parameters of the head-wise version with head dimension $H=32$.} Figure~\ref{fig:dynconv-scaling} (left) shows the validation loss as a function of compute (see Appendix~\ref{app:flop_calculations} for FLOP calculations). Fitting a curve to the results suggests that dynamic convolutions offer an approximate $1.33\times$ advantage over compute-matched Transformer baselines. 

We also experiment with applying dynamic convolutions to \emph{all linear layers}, instead of placing them only after the  $\texttt{qkv\_projection}$. To this end, we use the low-rank parameterization with rank $R=16$ across the $\{150M,300M,$ $600M,1B,2B\}$-parameter models. We find that placing dynamic short convolutions after every linear layer improves substantially over placing them on the queries, keys, and values alone. Fitting a scaling law for this variant (Figure~\ref{fig:dynconv-scaling} (right)) yields a $1.60 \times$ compute advantage over compute-matched Transformers, up from $1.33\times$ for the $Q$/$K$/$V$-only placement.

\begin{figure}[t] \centering \begin{minipage}{0.48\linewidth} \centering \includegraphics[width=\linewidth]{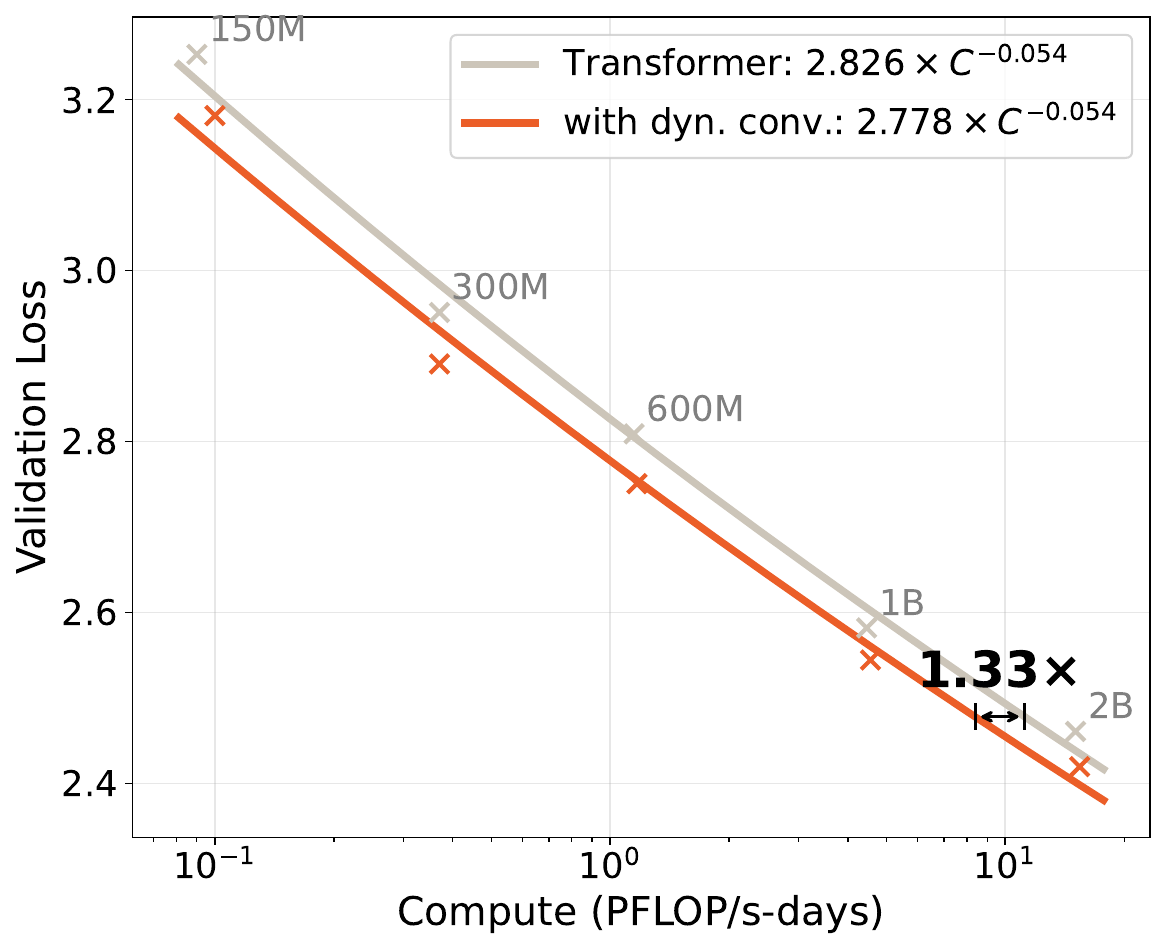} \end{minipage} \hfill \begin{minipage}{0.48\linewidth} \centering \includegraphics[width=\linewidth]{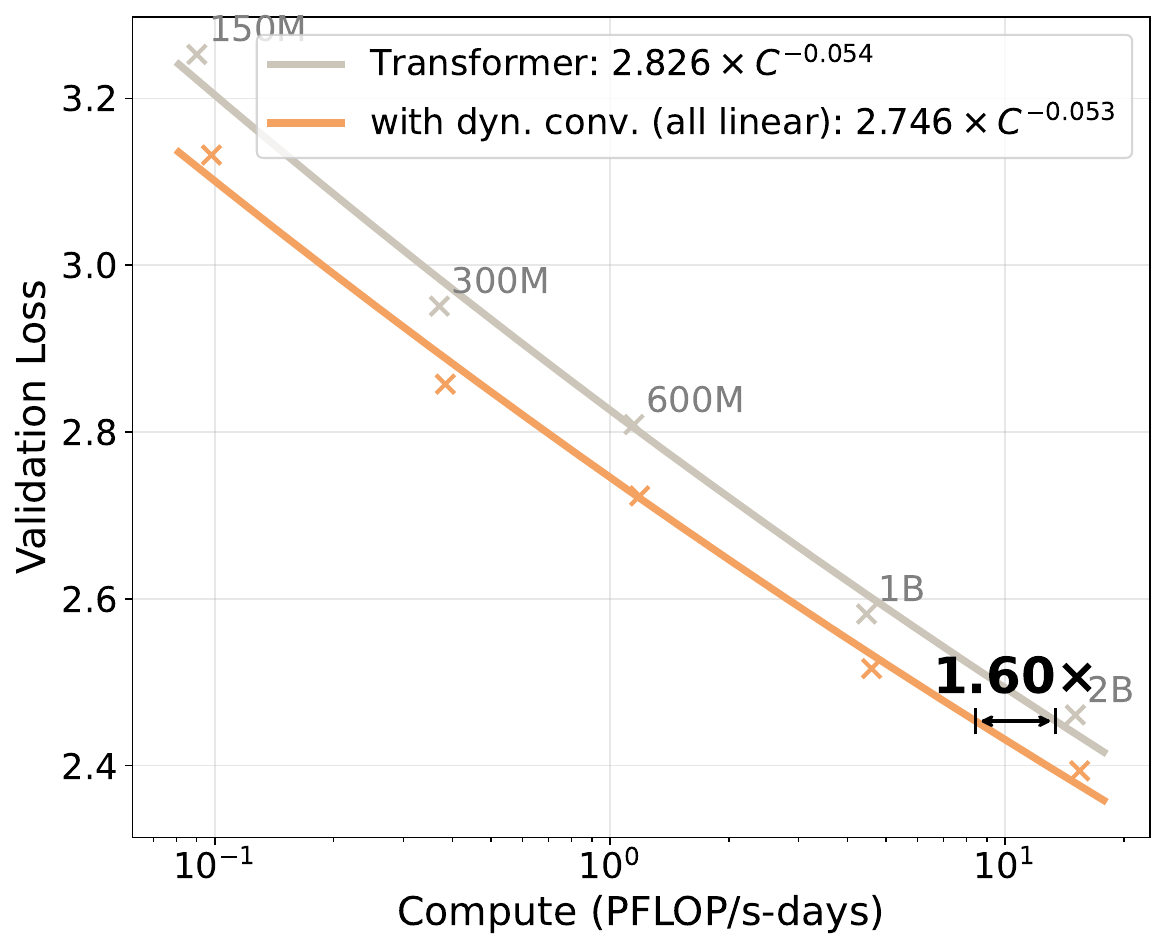} \end{minipage} \caption{Scaling laws on Transformers with low-rank dynamic convolutions applied to the keys, queries, and values (left) and placed after every linear layer (right).} \label{fig:dynconv-scaling} \end{figure}

\paragraph{Training throughput.} \begin{wrapfigure}{r}{0.56\linewidth}
    \centering
    \vspace{-4mm}
    \includegraphics[width=\linewidth]{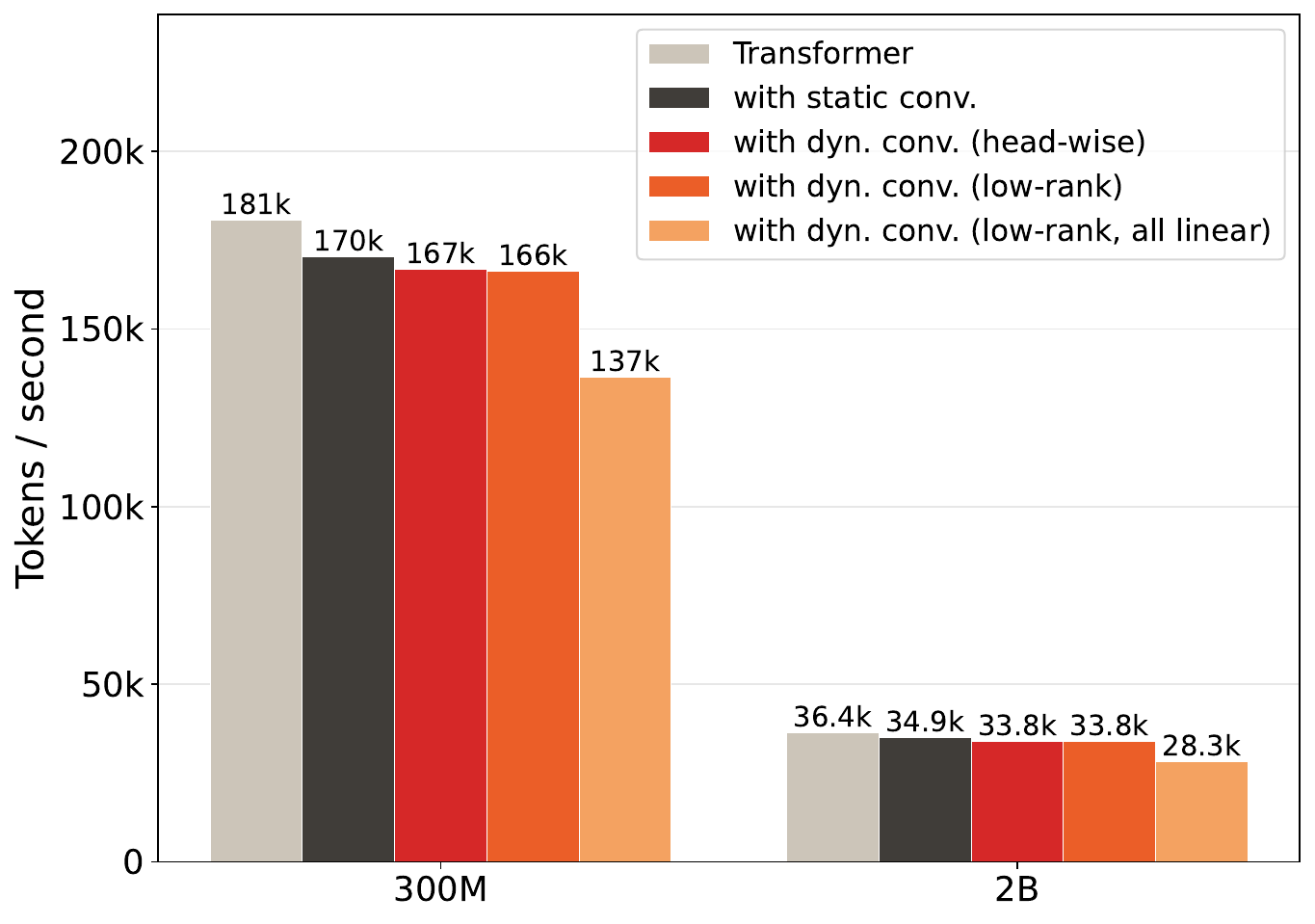}
    \caption{End-to-end training throughput measured on a single H100 80GB HBM3 GPU.}
    \vspace{-3mm}
    \label{fig:e2e-throughput}
\end{wrapfigure}
Figure~\ref{fig:e2e-throughput} shows that our dynamic convolution kernels are competitive with static convolution kernels. However, individual kernel efficiency does not always translate to end-to-end model efficiency.  We integrate our dynamic short-convolution kernels into the \texttt{lm-engine} codebase and measure end-to-end training throughput on a single H100 80GB HBM3 GPU. Figure~\ref{fig:e2e-throughput} reports tokens per second at the 300M and 2B parameter  scales (sequence length 4096, bf16 precision, with \texttt{torch.compile} enabled).  Both dynamic-convolution variants stay within 8\% overhead compared to the Transformer baseline across configurations, while adding static convolutions gives a slowdown of around 6\%.  We  expect more aggressive kernel fusion to close the gap for both static and dynamic variants. However, even with the current throughput penalty, our scaling law improvement ($1.33\times$) suggests a significant wall-clock time advantage.  

The all-linear variant results in a $22$--$25\%$ reduction in end-to-end training throughput. Future work could explore fusing the dynamic convolution into the matmul epilogue to further reduce memory I/O. This may substantially decrease this overhead and yield an even larger wall-clock time advantage.

\begin{table}[t]

\centering
\scriptsize
\setlength{\tabcolsep}{3.0pt}
\renewcommand{\arraystretch}{1.08}
\resizebox{\linewidth}{!}{
\begin{tabular}{l|ccc|ccc|c}
\toprule
\multirow{2}{*}{\textbf{Model}}
& \multirow{2}{*}{\makecell{\textbf{Train}\\\textbf{Tokens}}}
& \multirow{2}{*}{\textbf{Params}}
& \multirow{2}{*}{\makecell{\textbf{Conv.}\\\textbf{Location}}}
& \multicolumn{3}{c|}{\textbf{Perplexity}}
& \multicolumn{1}{c}{\textbf{Task}} \\
&
&
&
& \textbf{Nemo. $\downarrow$}
& \textbf{LAMB. $\downarrow$}
& \textbf{Wiki. $\downarrow$}
& \textbf{Avg. $\uparrow$} \\
\midrule
MoE Transformer
& \multirow{4}{*}{100B}
& 6.77B
& --
& 9.86 & 11.55 & 13.27
& 62.46 \\

\hspace{2mm} w/ static conv.
&
& 6.77B
& QKV
& 9.74 & 11.20 & 13.08
& 62.97 \\

\hspace{2mm} w/ dynamic conv. (head-wise)
&
& 6.80B
& QKV
& 9.65 & 11.61 & 12.90
& \textbf{63.43} \\

\hspace{2mm} w/ dynamic conv. (low-rank)
&
& 6.80B
& QKV
& \textbf{9.58} & \textbf{10.92} & \textbf{12.77}
& 63.42 \\

\midrule

Transformer
& \multirow{7}{*}{100B}
& 1.82B
& --
& 11.71 & 17.28 & 15.86
& 58.35 \\

\hspace{2mm} w/ more params (wider FFN)
&
& 1.87B
& --
& 11.67 & 18.49 & 15.72
& 58.23 \\

\hspace{2mm} w/ static conv.
&
& 1.83B
& QKV
& 11.50 & 16.21 & 15.43
& 58.94 \\

\hspace{2mm} w/ static conv. (param.\ matched)
&
& 1.88B
& QKV
& 11.44 & 15.61 & 15.33
& 58.97 \\

\hspace{2mm} w/ dynamic conv. (head-wise)
&
& 1.88B
& QKV
& 11.41 & 16.34 & 15.23
& 58.46 \\

\hspace{2mm} w/ dynamic conv. (low-rank)
&
& 1.88B
& QKV
& 11.24 & 15.43 & 14.98
& 59.70 \\

\hspace{2mm} w/ dynamic conv. (low-rank)
&
& 1.88B
& all linear
& \textbf{10.95} & \textbf{12.51} & \textbf{14.43}
& \textbf{60.70} \\

\midrule

Transformer
& \multirow{7}{*}{15B}
& 305.2M
& --
& 19.12 & 76.62 & 30.50
& 47.26 \\

\hspace{2mm} w/ more params (wider FFN)
&
& 311.5M
& --
& 18.99 & 68.05 & 30.04
& 46.64 \\

\hspace{2mm} w/ static conv.
&
& 305.4M
& QKV
& 18.66 & 69.64 & 29.50
& 46.86 \\

\hspace{2mm} w/ static conv. (param.\ matched)
&
& 311.8M
& QKV
& 18.63 & 66.53 & 29.37
& 46.22 \\

\hspace{2mm} w/ dynamic conv. (head-wise)
&
& 311.7M
& QKV
& 18.22 & 58.54 & 28.42
& 47.61 \\

\hspace{2mm} w/ dynamic conv. (low-rank)
&
& 311.8M
& QKV
& 18.01 & 56.66 & 27.98
& \textbf{48.90} \\

\hspace{2mm} w/ dynamic conv. (low-rank)
&
& 319.0M
& all linear
& \textbf{17.42} & \textbf{46.13} & \textbf{26.78}
& 48.81 \\

\midrule

Gated DeltaNet (w/o conv.)
& \multirow{5}{*}{15B}
& 305.2M
& --
& 18.93 & 69.90 & 30.24
& 46.93 \\

\hspace{2mm} w/ static conv.
&
& 305.4M
& QKV
& 18.75 & 67.63 & 29.82
& 46.76 \\

\hspace{2mm} w/ static conv. (param.\ matched)
&
& 309.6M
& QKV
& 18.74 & 69.06 & 29.78
& 47.60 \\

\hspace{2mm} w/ dynamic conv. (head-wise)
&
& 309.6M
& QKV
& 18.03 & 59.49 & 28.17
& 47.27 \\

\hspace{2mm} w/ dynamic conv. (low-rank)
&
& 309.5M
& QKV
& \textbf{17.95} & \textbf{50.56} & \textbf{27.95}
& \textbf{49.22} \\

\midrule

Mamba-2 (w/o conv.)
& \multirow{5}{*}{15B}
& 306.2M
& --
& 20.26 & 80.81 & 33.26
& 46.41 \\

\hspace{2mm} w/ static conv.
&
& 306.4M
& QKV
& 19.30 & 83.50 & 31.32
& 45.78 \\

\hspace{2mm} w/ static conv. (param.\ matched)
&
& 309.8M
& QKV
& 19.26 & 78.96 & 31.17
& 46.78 \\

\hspace{2mm} w/ dynamic conv. (head-wise)
&
& 309.8M
& QKV
& \textbf{18.69} & \textbf{65.80} & 30.03
& 47.24 \\

\hspace{2mm} w/ dynamic conv. (low-rank)
&
& 309.8M
& QKV
& 18.72 & 71.12 & \textbf{29.90}
& \textbf{47.34} \\

\bottomrule
\end{tabular}
}
\vspace{4mm}
\caption{Main evaluation results of Transformers, Transformers with Static Short Convolutions, and Transformers with Dynamic Short Convolutions. \emph{Param.\ matched} static baselines widen the MLP intermediate dimension to exactly match the parameter count of the dynamic variants. \emph{Task Avg.} averages 0-shot accuracy over 11 lm-eval-harness tasks.}
\label{tab:main_eval_compact}
\vspace{-3mm}
\end{table}

\paragraph{Evaluation.} We next perform downstream evaluations across several baselines. First, Transformer uses the same training recipe and architecture, but without any convolutional layers. Second, Transformer (more params) increases the MLP intermediate dimension to account for the additional parameters introduced through our dynamic convolutions.
Third, we compare against a Transformer with static short convolutions on top of the queries, keys, and values, following the ``Canon-B'' setup from \citet{allen2025physics}. Finally, we compare dynamic short convolutions with both low-rank and head-wise parameterizations.
We report perplexity on 25M tokens of held-out Nemotron-CC data, as well as on Wikitext-103 \citep{merity2016pointer} and LAMBADA \citep{paperno_lambada_2016}. Additionally, we report zero-shot accuracy on various common-sense reasoning tasks via \texttt{lm-evaluation-harness} \citep{eval-harness}.

Table~\ref{tab:main_eval_compact} reports results for models with (roughly) 300M (15B tokens), 2B (100B tokens), and 7B/1B active Mixture of Experts (100B tokens). Static convolutions generally improve upon ordinary Transformers, but both the head-wise and low-rank variants outperform static convolutions on perplexity and task accuracy at all parameter scales. The all-linear variant gives further gains. See Table~\ref{tab:full_0shot} of the Appendix for the task performance broken down by benchmark.

We now analyze the capabilities of these models for in-context learning and retrieval on the RULER benchmark \citep{hsieh2024ruler}. 
RULER is generally a difficult benchmark for models trained at our scale. We analyze the RULER results for the strongest MoE models in Table~\ref{tab:7b_ruler_4k}. We find that the dynamic convolutions perform particularly well on the multi-key (MK), multi-query (MQ), and multi-value (MV) subtasks within RULER, which makes sense given the ability to perform input-dependent local aggregations enabled by the dynamic convolutions. See Table~\ref{tab:full_ruler_4k} of the Appendix for the RULER results for all models. 


\begin{table}[t]
\centering
\scriptsize
\setlength{\tabcolsep}{2.5pt}
\renewcommand{\arraystretch}{1.05}
\caption{Per-subtask RULER accuracy (\%) at context length 4096.}
\label{tab:7b_ruler_4k}
\resizebox{\linewidth}{!}{
\begin{tabular}{l|ccccccccccc|c}
\toprule
\textbf{Model} & \textbf{S1} & \textbf{S2} & \textbf{S3} & \textbf{MK1} & \textbf{MK2} & \textbf{MK3} & \textbf{MQ} & \textbf{MV} & \textbf{CWE} & \textbf{FWE} & \textbf{VT} & \textbf{Avg.} \\
\midrule

MoE Transformer &  99.8 & \textbf{100.0} & 83.8 & 70.0 & 4.8 & 21.2 & 39.4 & 32.0 & 26.4 & \textbf{43.3} & 9.3 & 48.2 \\
w/ static conv. &  \textbf{100.0} & \textbf{100.0} & 73.8 & 72.0 & 27.6 & 6.0 & 36.5 & 40.8 & 13.3 & 26.8 & \textbf{23.9} & 47.3 \\
w/ dynamic conv. (head-wise) &  \textbf{100.0} & 99.8 & 81.8 & \textbf{74.2} & \textbf{45.4} & \textbf{38.4} & 37.4 & 40.5 & 15.1 & 16.3 & 8.1 & 50.6 \\
w/ dynamic conv. (low-rank) &  99.8 & \textbf{100.0} & \textbf{93.0} & 66.2 & 12.8 & 11.8 & \textbf{48.2} & \textbf{50.4} & \textbf{31.1} & 26.4 & 18.4 & \textbf{50.7} \\
\bottomrule
\end{tabular}
}
\vspace{-0mm}
\end{table}

\paragraph{Linear attention variants.}
Modern linear RNN architectures such as Mamba and DeltaNet already include static depthwise separable short convolutions on $Q$, $K$, $V$ as part of their sequence mixer \citep{Gu2023MambaLS, dao2024transformers, yang2024gated}. We test whether replacing these static convolutions with our dynamic convolutions further improves the architecture. At the bottom of Table~\ref{tab:main_eval_compact} we report the  300M/15B-token results for Mamba-2 \citep{dao2024transformers} and Gated DeltaNet \citep{schlag_linear_2021, yang2024parallelizing, yang2024gated}: the default (with static conv.), without any convolutions, and with  two variants of our dynamic short convolutions. 
As expected, removing the static short convolutions increases perplexity on held-out training data. Replacing the static convolutions with their dynamic counterparts significantly improves performance in terms of perplexity across all datasets. Notably, Mamba-2 with dynamic convolutions performs about as well as Gated DeltaNet with static convolutions, which suggests that incorporating dynamic short convolutions may be more beneficial than redesigning the sequence mixer.

\vspace{2mm}
\subsection{Ablations}
\label{sec:ablations}
We next perform a series of ablations on our architectural decisions. Here we work with the 300M models trained on 15B tokens, and report perplexity (PPL) on the Nemotron-CC corpus.

\paragraph{Width, head dimension, rank.} Our main experiments used filter width $W$ of 4, head-size $H$ of 32, and for the low-rank version, rank $R$ such that it is param-matched to the head-wise version. We perform a sweep of these hyperparameters. Table~\ref{tab:ablation_sweep} shows the ablation results.

For width, we find that 3 or 4 is generally the sweet spot in terms of providing the best performance. This has generally been found to be the case for static convolutions as well. Widths beyond this sweet spot do not provide additional gains even though they add parameters. For head dimension in the head-wise variant, we used $H=32$ for our experiments. Making the head dimension smaller does improve performance, but results in many additional parameters. For the low-rank variant, increasing the expressivity via increasing $R$ unsurprisingly leads to improved performance, but at the cost of more parameters. Dynamic convolutions therefore provide another axis with which to trade off compute/parameters for performance. Overall, the results suggest that the low-rank parameterization with $R=16$ offers a strong trade-off between performance and parameter count.

\begin{table}[!t]
\centering
\small
\setlength{\tabcolsep}{4pt}
\caption{Ablations on the 300M models trained on 15B tokens, reporting Nemotron-CC perplexity. (a) Sweep over kernel width $W$, head size $H$, and rank $R$ for dynamic convolutions on Q+K+V. (b) Placement of dynamic convolutions inside the attention layer (low-rank $R{=}16$, $W{=}4$). (c) QK-norm Transformers with and without convolutions.}
\label{tab:ablations}
\begin{minipage}[t]{0.49\textwidth}
    \begin{subtable}[t]{\linewidth}
        \centering
        \caption{Width / head size / rank sweep.}
        \label{tab:ablation_sweep}
        \begin{tabular}{lcc}
        \toprule
        \textbf{Sweep} & \textbf{Params} & \textbf{PPL} \\
        \midrule
        \multicolumn{3}{l}{\emph{Width $W$ (low-rank, $R{=}16$)}} \\
        \hspace{2mm} $W=1$  & 306.8M & 18.42 \\
        \hspace{2mm} $W=2$  & 307.6M & 18.17 \\
        \hspace{2mm} $W=3$  & 308.5M & \textbf{18.08} \\
        \hspace{2mm} $W=4$  & 309.3M & 18.10 \\
        \hspace{2mm} $W=5$  & 310.1M & 18.09 \\
        \hspace{2mm} $W=6$  & 311.0M & 18.10 \\
        \midrule
        \multicolumn{3}{l}{\emph{Head size $H$ (head-wise, $W{=}4$)}} \\
        \hspace{2mm} $H=8$    & 330.5M & \textbf{18.03} \\
        \hspace{2mm} $H=16$   & 317.9M & 18.08 \\
        \hspace{2mm} $H=32$   & 311.7M & 18.21 \\
        \hspace{2mm} $H=64$   & 308.5M & 18.25 \\
        \hspace{2mm} $H=128$  & 306.9M & 18.40 \\
        \midrule
        \multicolumn{3}{l}{\emph{Rank $R$ (low-rank, $W{=}4$)}} \\
        \hspace{2mm} $R=4$    & 306.3M & 18.26 \\
        \hspace{2mm} $R=8$    & 307.3M & 18.19 \\
        \hspace{2mm} $R=16$   & 309.3M & 18.10 \\
        \hspace{2mm} $R=32$   & 313.2M & 18.04 \\
        \hspace{2mm} $R=64$   & 321.1M & 17.87 \\
        \hspace{2mm} $R=128$  & 336.8M & \textbf{17.85} \\
        \bottomrule
        \end{tabular}
    \end{subtable}
\end{minipage}\hfill
\begin{minipage}[t]{0.49\textwidth}
    \begin{subtable}[t]{\linewidth}
        \centering
        \caption{Layer placement.}
        \label{tab:ablation_placement}
        \begin{tabular}{lcc}
        \toprule
        \textbf{Placement} & \textbf{Params} & \textbf{PPL} \\
        \midrule
        Transformer (w/o conv.) & 305.2M & 19.12 \\
        Q only          & 306.5M & 18.69 \\
        K only          & 306.5M & 18.83 \\
        V only          & 306.5M & 18.56 \\
        Q + K           & 307.9M & 18.44 \\
        Q + V           & 307.9M & 18.36 \\
        K + V           & 307.9M & 18.35 \\
        Q + K + V       & 309.3M & \textbf{18.10} \\
        \bottomrule
        \end{tabular}
    \end{subtable}

    \vspace{2.5em}

    \begin{subtable}[t]{\linewidth}
        \centering
        \caption{QK-norm Transformers.}
        \label{tab:ablation_qknorm}
        \begin{tabular}{lcc}
        \toprule
        \textbf{Setup} & \textbf{Params} & \textbf{PPL} \\
        \midrule
        Transformer with QK-Norm     & 305.2M & 18.69 \\
        w/ static conv.              & 305.4M & 18.56 \\
        w/ dynamic conv. (head-wise) & 311.5M & 18.30 \\
        w/ dynamic conv. (low-rank)  & 311.6M & \textbf{17.95} \\
        \bottomrule
        \end{tabular}
    \end{subtable}
\end{minipage}
\end{table}

\paragraph{Layer placement.} In our main experiments we placed dynamic convolutions on all components of attention, i.e., queries, keys, and values. We ablate this design choice by placing dynamic convolutions on different subsets of QKV. Table~\ref{tab:ablation_placement} shows that the largest single-projection placement gain comes from the value projection. Applying them to two projections improves performance further, while the best result is achieved when dynamic convolutions are applied to all three projections. We therefore use Q+K+V placement in our main experiments.

\paragraph{QK-norm Transformers.} Our main experiments were conducted on Transformers without QK-norm. However, QK-norm \citep{pmlr-v202-dehghani23a} is becoming a popular part of recent frontier open-source LLMs \citep{5team2025glm45agenticreasoningcoding, yang2025qwen3technicalreport, gemmateam2025gemma3technicalreport}. Would dynamic convolutions be helpful for Transformers trained with QK-norm? We show the results on QK-norm Transformers in Table~\ref{tab:ablation_qknorm}, where we indeed find that dynamic convolutions continue to provide significant gains for Transformers trained with QK-norm. By contrast, static convolutions provide little benefit when combined with QK-norm in our experiments. For these experiments, we apply per-head RMSNorm to the queries and keys after the convolution (when used) and before RoPE.

\section{Discussion and Limitations}

Our results suggest that dynamic short convolutions are a useful primitive for improving Transformer-based language models. Unlike static short convolutions, which impose the same local aggregation rule at every position, dynamic convolutions allow each token to choose an input-dependent local composition function. The synthetic experiments support the utility of such a layer, showing gains on tasks that require resolving variable-length local structure before performing recall. The language-modeling results further suggest that these benefits are not limited to toy settings: dynamic convolutions improve dense Transformers, MoE Transformers, and linear attention variants, and the improvements persist under parameter-matched and compute-matched comparisons.

On the limitations side, our scaling study reaches 2B dense parameters and a 7B-parameter MoE with 1B active parameters, which is sufficient to establish consistent trends but does not by itself demonstrate that the same compute advantage will hold at frontier scale, under substantially longer training, or across different data mixtures and tokenizers. Moreover, while our Triton implementations make dynamic convolutions practical on H100 GPUs, additional engineering would be needed to fully optimize inference, support a broader range of hardware, and fuse the dynamic and static components more aggressively. Finally, we only explore a small subset of possible parameterizations, placements, and kernel widths.

\section{Related Work}

\paragraph{Convolutional networks for sequence modeling.}
Convolutional networks were a popular  class of  neural sequence models before the rise of attention-based architectures. Early neural NLP systems used one-dimensional convolutions over word embeddings for tagging and sentence classification \citep{collobert2008unified,collobert2011nlp,kalchbrenner2014convolutional,kim2014convolutional}. Later work scaled convolutional sequence models to machine translation and language modeling using dilations, gating, and stacked convolutional blocks \citep{kalchbrenner2016neural,gehring2017convolutional,dauphin2017language,bai2018tcn}. More recently, long-convolution models such as S4 and Hyena have revisited convolutions as subquadratic alternatives to attention by using implicit long filters and gating \citep{gu2022efficiently,poli2023hyena}. 

\paragraph{Dynamic convolutions.}
Dynamic convolutions have a long history in vision models \citep{jia2016dynamic,yang2019condconv,li2019selective,chen2020dynamicconv,zhou2021decoupled,li2022omni}. Their use in language processing and sequence modeling has been more limited. \citet{wu2019pay} introduced lightweight and dynamic convolutions that predict convolution kernels from the current time step as efficient alternatives to self-attention, and ConvBERT uses span-based dynamic convolutions to replace a subset of BERT attention heads for local dependency modeling \citep{jiang2020convbert}.

\paragraph{Convolutions in modern Transformers and linear RNNs.}
Since the introduction of Transformers, there have been works that combine attention with static convolutions. Conformer combines self-attention with convolutional modules for speech recognition \citep{gulati2020conformer}, and Lite Transformer allocates separate branches to long-range attention and short-range convolution \citep{wu2020lite}. In language modeling, Primer found through architecture search that adding depthwise convolutions after the query, key, and value projections substantially improves training efficiency \citep{so2021primer}; more recent work emphasizes horizontal information flow among neighboring tokens across multiple sequence architectures \citep{allen2025physics}. Static short convolutions are also standard in recent linear RNNs \citep{Gu2023MambaLS,dao2024transformers,yang2024parallelizing,yang2024gated}. Recent work \citep{gu2026jet}  uses dynamic convolutions just in the value layer when converting pretrained softmax attention layers  to linear  attention layers. Our results show that dynamic convolutions improve upon static convolutions in both Transformers and linear RNNs when pretrained from scratch.

\section{Conclusion}
We introduced dynamic short convolutions as an input-dependent, locality-biased primitive for improving Transformer-based language models. By generating convolutional filters from the current hidden state, dynamic convolutions extend static short convolutions with greater expressivity while retaining efficient local sequence mixing. Across synthetic associative-recall tasks, dense language-modeling experiments from 150M to 2B parameters, a 7B-parameter MoE model, and two linear attention architectures, dynamic convolutions consistently improve over both standard Transformers and static-convolution baselines. Our scaling-law analysis indicates a meaningful compute advantage, and our custom Triton kernels show that these gains can be obtained with modest end-to-end training overhead. Taken together, these results suggest that dynamic short convolutions are a scalable and practical architectural primitive, complementary to attention and promising for LLMs.



\begin{ack}
We would like to thank Han Guo, Assaf Ben-Kish, and Yanick Schimpf
for valuable discussions and feedback.

This study was supported by MIT-IBM Watson AI Lab and the AI2050
program at Schmidt Sciences (Grant G-25-67980).

\textbf{Competing interests.}
The authors declare no competing financial interests.
\end{ack}

\bibliography{references}

@article{gu2026jet,
  title={Jet-nemotron: Efficient language model with post neural architecture search},
  author={Gu, Yuxian and Hu, Qinghao and Xi, Haocheng and Chen, Junyu and Yang, Shang and Han, Song and Cai, Han},
  journal={Advances in Neural Information Processing Systems},
  volume={38},
  pages={47191--47218},
  year={2026}
}

@article{li2022omni,
  title={Omni-dimensional dynamic convolution},
  author={Li, Chao and Zhou, Aojun and Yao, Anbang},
  journal={arXiv preprint arXiv:2209.07947},
  year={2022}
}

@inproceedings{chollet2017xception,
  title={Xception: Deep learning with depthwise separable convolutions},
  author={Chollet, Fran{\c{c}}ois},
  booktitle={Proceedings of the IEEE conference on computer vision and pattern recognition},
  pages={1251--1258},
  year={2017}
}

@article{hsieh2024ruler,
  title={RULER: What's the real context size of your long-context language models?},
  author={Hsieh, Cheng-Ping and Sun, Simeng and Kriman, Samuel and Acharya, Shantanu and Rekesh, Dima and Jia, Fei and Zhang, Yang and Ginsburg, Boris},
  booktitle={Proceedings of COLM},
  year={2024}
}

@article{howard2017mobilenets,
  title={Mobilenets: Efficient convolutional neural networks for mobile vision applications},
  author={Howard, Andrew G and Zhu, Menglong and Chen, Bo and Kalenichenko, Dmitry and Wang, Weijun and Weyand, Tobias and Andreetto, Marco and Adam, Hartwig},
  journal={arXiv preprint arXiv:1704.04861},
  year={2017}
}

@article{lahoti2026mamba,
  title={Mamba-3: Improved sequence modeling using state space principles},
  author={Lahoti, Aakash and Li, Kevin Y and Chen, Berlin and Wang, Caitlin and Bick, Aviv and Kolter, J Zico and Dao, Tri and Gu, Albert},
  booktitle={Proceedings of ICLR},
  year={2026}
}

@article{afzalbick2026raven,
  title={Raven: High-Recall Sequence Modeling with Sparse Memory Routing},
  author={Afzal, Arshia and Bick, Aviv and Xing, Eric P. and Cevher, Volkan and Gu, Albert},
  year={2026},
  publisher={MDPI}
}

@inproceedings{zhou2021decoupled,
  title={Decoupled dynamic filter networks},
  author={Zhou, Jingkai and Jampani, Varun and Pi, Zhixiong and Liu, Qiong and Yang, Ming-Hsuan},
  booktitle={Proceedings of the IEEE/CVF conference on computer vision and pattern recognition},
  pages={6647--6656},
  year={2021}
}

@inproceedings{li2019selective,
  title={Selective kernel networks},
  author={Li, Xiang and Wang, Wenhai and Hu, Xiaolin and Yang, Jian},
  booktitle={Proceedings of the IEEE/CVF conference on computer vision and pattern recognition},
  pages={510--519},
  year={2019}
}

@inproceedings{jia2016dynamic,
  title     = {Dynamic Filter Networks},
  author    = {Jia, Xu and De Brabandere, Bert and Tuytelaars, Tinne and Van Gool, Luc},
  booktitle = {Advances in Neural Information Processing Systems},
  volume    = {29},
  pages     = {667--675},
  year      = {2016}
}

@article{bai2018tcn,
  title   = {An Empirical Evaluation of Generic Convolutional and Recurrent Networks for Sequence Modeling},
  author  = {Bai, Shaojie and Kolter, J. Zico and Koltun, Vladlen},
  journal = {arXiv preprint arXiv:1803.01271},
  year    = {2018}
}

@inproceedings{yang2019condconv,
  title     = {CondConv: Conditionally Parameterized Convolutions for Efficient Inference},
  author    = {Yang, Brandon and Bender, Gabriel and Le, Quoc V. and Ngiam, Jiquan},
  booktitle = {Advances in Neural Information Processing Systems},
  year      = {2019}
}

@inproceedings{chen2020dynamicconv,
  title     = {Dynamic Convolution: Attention over Convolution Kernels},
  author    = {Chen, Yinpeng and Dai, Xiyang and Liu, Mengchen and Chen, Dongdong and Yuan, Lu and Liu, Zicheng},
  booktitle = {Proceedings of the IEEE/CVF Conference on Computer Vision and Pattern Recognition},
  year      = {2020}
}

@inproceedings{wu2019pay,
  title     = {Pay Less Attention with Lightweight and Dynamic Convolutions},
  author    = {Wu, Felix and Fan, Angela and Baevski, Alexei and Dauphin, Yann N. and Auli, Michael},
  booktitle = {International Conference on Learning Representations},
  year      = {2019},
}

@misc{merity2016pointer,
      title={Pointer Sentinel Mixture Models},
      author={Stephen Merity and Caiming Xiong and James Bradbury and Richard Socher},
      year={2016},
      eprint={1609.07843},
      archivePrefix={arXiv},
      primaryClass={cs.CL}
}

@inproceedings{jiang2020convbert,
  title     = {ConvBERT: Improving BERT with Span-based Dynamic Convolution},
  author    = {Jiang, ZiHang and Yu, Weihao and Zhou, Daquan and Chen, Yunpeng and Feng, Jiashi and Yan, Shuicheng},
  booktitle = {Advances in Neural Information Processing Systems},
  year      = {2020}
}

@inproceedings{gulati2020conformer,
  title     = {Conformer: Convolution-augmented Transformer for Speech Recognition},
  author    = {Gulati, Anmol and Qin, James and Chiu, Chung-Cheng and Parmar, Niki and Zhang, Yu and Yu, Jiahui and Han, Wei and Wang, Shibo and Zhang, Zhengdong and Wu, Yonghui and Pang, Ruoming},
  booktitle = {Interspeech},
  year      = {2020}
}

@inproceedings{wu2020lite,
  title     = {Lite Transformer with Long-Short Range Attention},
  author    = {Wu, Zhanghao and Liu, Zhijian and Lin, Ji and Lin, Yujun and Han, Song},
  booktitle = {International Conference on Learning Representations},
  year      = {2020}
}

@inproceedings{dao2024transformers,
  title     = {Transformers are {SSM}s: Generalized Models and Efficient Algorithms Through Structured State Space Duality},
  author    = {Dao, Tri and Gu, Albert},
  booktitle = {Proceedings of the 41st International Conference on Machine Learning},
  pages     = {10041--10071},
  year      = {2024},
  volume    = {235},
  series    = {Proceedings of Machine Learning Research},
  publisher = {PMLR}
}

@inproceedings{paszke2019pytorch,
  title     = {PyTorch: An Imperative Style, High-Performance Deep Learning Library},
  author    = {Paszke, Adam and Gross, Sam and Massa, Francisco and Lerer, Adam and Bradbury, James and Chanan, Gregory and Killeen, Trevor and Lin, Zeming and Gimelshein, Natalia and Antiga, Luca and Desmaison, Alban and Kopf, Andreas and Yang, Edward and DeVito, Zachary and Raison, Martin and Tejani, Alykhan and Chilamkurthy, Sasank and Steiner, Benoit and Fang, Lu and Bai, Junjie and Chintala, Soumith},
  booktitle = {Advances in Neural Information Processing Systems},
  volume    = {32},
  year      = {2019}
}

@inproceedings{shazeer2017outrageously,
  title     = {Outrageously Large Neural Networks: The Sparsely-Gated Mixture-of-Experts Layer},
  author    = {Shazeer, Noam and Mirhoseini, Azalia and Maziarz, Krzysztof and Davis, Andy and Le, Quoc V. and Hinton, Geoffrey E. and Dean, Jeff},
  booktitle = {International Conference on Learning Representations},
  year      = {2017},
  url       = {https://arxiv.org/abs/1701.06538}
}

@misc{loshchilov2019decoupledweightdecayregularization,
      title={Decoupled Weight Decay Regularization}, 
      author={Ilya Loshchilov and Frank Hutter},
      year={2019},
      eprint={1711.05101},
      archivePrefix={arXiv},
      primaryClass={cs.LG},
      url={https://arxiv.org/abs/1711.05101}, 
}

@inproceedings{su-etal-2025-nemotron,
    title = "Nemotron-{CC}: Transforming {C}ommon {C}rawl into a Refined Long-Horizon Pretraining Dataset",
    author = "Su, Dan  and
      Kong, Kezhi  and
      Lin, Ying  and
      Jennings, Joseph  and
      Norick, Brandon  and
      Kliegl, Markus  and
      Patwary, Mostofa  and
      Shoeybi, Mohammad  and
      Catanzaro, Bryan",
    editor = "Che, Wanxiang  and
      Nabende, Joyce  and
      Shutova, Ekaterina  and
      Pilehvar, Mohammad Taher",
    booktitle = "Proceedings of the 63rd Annual Meeting of the Association for Computational Linguistics (Volume 1: Long Papers)",
    month = jul,
    year = "2025",
    address = "Vienna, Austria",
    publisher = "Association for Computational Linguistics",
    url = "https://aclanthology.org/2025.acl-long.123/",
    doi = "10.18653/v1/2025.acl-long.123",
    pages = "2459--2475",
    ISBN = "979-8-89176-251-0"
}

@software{mishra2024lmengine,
  title = {LM Engine: A Hyper-Optimized Library for Pretraining and Finetuning},
  author = {Mishra, Mayank},
  year = {2024},
  url = {https://github.com/open-lm-engine/lm-engine}
}

@inproceedings{ansel2024pytorch2,
  title     = {PyTorch 2: Faster Machine Learning Through Dynamic Python Bytecode Transformation and Graph Compilation},
  author    = {Ansel, Jason and Yang, Edward and He, Horace and Gimelshein, Natalia and Jain, Animesh and Voznesensky, Michael and Bao, Bin and Bell, Peter and Berard, David and Burovski, Evgeni and Chauhan, Geeta and Chourdia, Anjali and Constable, Will and Desmaison, Alban and DeVito, Zachary and Ellison, Elias and Feng, Will and Gong, Jiong and Gschwind, Michael and Hirsh, Brian and Huang, Sherlock and Kalambarkar, Kshiteej and Kirsch, Laurent and Lazos, Michael and Lezcano, Mario and Liang, Yanbo and Liang, Jason and Lu, Yinghai and Luk, C. K. and Maher, Bert and Pan, Yunjie and Puhrsch, Christian and Reso, Matthias and Saroufim, Mark and Siraichi, Marcos Yukio and Suk, Helen and Zhang, Shunting and Suo, Michael and Tillet, Phil and Zhao, Xu and Wang, Eikan and Zhou, Keren and Zou, Richard and Wang, Xiaodong and Mathews, Ajit and Wen, William and Chanan, Gregory and Wu, Peng and Chintala, Soumith},
  booktitle = {Proceedings of the 29th ACM International Conference on Architectural Support for Programming Languages and Operating Systems, Volume 2},
  pages     = {929--947},
  year      = {2024},
  publisher = {Association for Computing Machinery},
  doi       = {10.1145/3620665.3640366}
}

@article{allen2025physics,
  title={Physics of Language Models: Part 4.1, Architecture design and the magic of Canon layers},
  author={Allen-Zhu, Zeyuan},
  journal={arXiv preprint arXiv:2512.17351},
  year={2025}
}

@inproceedings{so2021primer,
  title={Primer: Searching for Efficient Transformers for Language Modeling},
  author={So, David and M{\"a}nke, Wojciech and Liu, Hanxiao and Dai, Zihang and Shazeer, Noam and Le, Quoc V},
  booktitle={Advances in Neural Information Processing Systems},
  volume={34},
  pages={26053--26066},
  year={2021}
}

@inproceedings{dauphin2017language,
  title={Language modeling with gated convolutional networks},
  author={Dauphin, Yann N and Fan, Angela and Auli, Michael and Grangier, David},
  booktitle={International conference on machine learning},
  pages={933--941},
  year={2017},
  organization={PMLR}
}

@inproceedings{poli2023hyena,
  title={Hyena hierarchy: Towards larger convolutional language models},
  author={Poli, Michael and Massaroli, Stefano and Nguyen, Eric and Fu, Daniel Y and Dao, Tri and Baccus, Stephen and Bengio, Yoshua and Ermon, Stefano and R{\'e}, Christopher},
  booktitle={International Conference on Machine Learning},
  pages={28043--28078},
  year={2023},
  organization={PMLR}
}

@article{collobert2011nlp,
  title   = {Natural Language Processing (Almost) from Scratch},
  author  = {Collobert, Ronan and Weston, Jason and Bottou, L{\'e}on and Karlen, Michael and Kavukcuoglu, Koray and Kuksa, Pavel},
  journal = {Journal of Machine Learning Research},
  year    = {2011}
}

@inproceedings{kim2014convolutional,
  title={Convolutional neural networks for sentence classification},
  author={Kim, Yoon},
  booktitle={Proceedings of the 2014 conference on empirical methods in natural language processing (EMNLP)},
  pages={1746--1751},
  year={2014}
}

@inproceedings{kalchbrenner2014convolutional,
  title={A convolutional neural network for modelling sentences},
  author={Kalchbrenner, Nal and Grefenstette, Edward and Blunsom, Phil},
  booktitle={Proceedings of the 52nd Annual Meeting of the Association for Computational Linguistics (Volume 1: Long Papers)},
  pages={655--665},
  year={2014}
}

@inproceedings{collobert2008unified,
  title={A unified architecture for natural language processing: Deep neural networks with multitask learning},
  author={Collobert, Ronan and Weston, Jason},
  booktitle={Proceedings of the 25th international conference on Machine learning},
  pages={160--167},
  year={2008}
}

@inproceedings{gehring2017convolutional,
  title={Convolutional sequence to sequence learning},
  author={Gehring, Jonas and Auli, Michael and Grangier, David and Yarats, Denis and Dauphin, Yann N},
  booktitle={International conference on machine learning},
  pages={1243--1252},
  year={2017},
  organization={PMLR}
}

@article{kalchbrenner2016neural,
  title={Neural machine translation in linear time},
  author={Kalchbrenner, Nal and Espeholt, Lasse and Simonyan, Karen and Oord, Aaron van den and Graves, Alex and Kavukcuoglu, Koray},
  journal={arXiv preprint arXiv:1610.10099},
  year={2016}
}

@inproceedings{xiong2020layernorm,
  author    = {Xiong, Ruibin and Yang, Yunchang and He, Di and Zheng, Kai and Zheng, Shuxin and Xing, Chen and Zhang, Huishuai and Lan, Yanyan and Wang, Liwei and Liu, Tie-Yan},
  title     = {On Layer Normalization in the Transformer Architecture},
  booktitle = {Proceedings of the 37th International Conference on Machine Learning},
  series    = {Proceedings of Machine Learning Research},
  volume    = {119},
  pages     = {10524--10533},
  year      = {2020},
}

@inproceedings{zhang2019rmsnorm,
  author    = {Zhang, Biao and Sennrich, Rico},
  title     = {Root Mean Square Layer Normalization},
  booktitle = {Advances in Neural Information Processing Systems 32},
  year      = {2019},
}

@article{su2021roformer,
  author        = {Su, Jianlin and Lu, Yu and Pan, Shengfeng and Wen, Bo and Liu, Yunfeng},
  title         = {RoFormer: Enhanced Transformer with Rotary Position Embedding},
  journal       = {arXiv preprint arXiv:2104.09864},
  year          = {2021},
  eprint        = {2104.09864},
}

@article{shazeer2019mqa,
  author        = {Shazeer, Noam},
  title         = {Fast Transformer Decoding: One Write-Head is All You Need},
  journal       = {arXiv preprint arXiv:1911.02150},
  year          = {2019},
  eprint        = {1911.02150},
}

@inproceedings{ainslie2023gqa,
  author    = {Ainslie, Joshua and Lee-Thorp, James and de Jong, Michiel and Zemlyanskiy, Yury and Lebr{\'o}n, Federico and Sanghai, Sumit},
  title     = {GQA: Training Generalized Multi-Query Transformer Models from Multi-Head Checkpoints},
  booktitle = {Proceedings of the 2023 Conference on Empirical Methods in Natural Language Processing},
  year      = {2023},
}

@article{fedus2022switch,
  author  = {Fedus, William and Zoph, Barret and Shazeer, Noam},
  title   = {Switch Transformers: Scaling to Trillion Parameter Models with Simple and Efficient Sparsity},
  journal = {Journal of Machine Learning Research},
  year    = {2022},
}

@article{shazeer2020glu,
  author        = {Shazeer, Noam},
  title         = {GLU Variants Improve Transformer},
  journal       = {arXiv preprint arXiv:2002.05202},
  year          = {2020},
  url           = {https://arxiv.org/abs/2002.05202}
}

@inproceedings{ioffe2015batchnorm,
  author    = {Ioffe, Sergey and Szegedy, Christian},
  title     = {Batch Normalization: Accelerating Deep Network Training by Reducing Internal Covariate Shift},
  booktitle = {Proceedings of the 32nd International Conference on Machine Learning},
  series    = {Proceedings of Machine Learning Research},
  volume    = {37},
  pages     = {448--456},
  year      = {2015},
  publisher = {PMLR},
  url       = {https://proceedings.mlr.press/v37/ioffe15.html}
}

@inproceedings{he2016deep,
  title={Deep residual learning for image recognition},
  author={He, Kaiming and Zhang, Xiangyu and Ren, Shaoqing and Sun, Jian},
  booktitle={Proceedings of the IEEE conference on computer vision and pattern recognition},
  pages={770--778},
  year={2016}
}

@article{ba2016layernorm,
  author        = {Ba, Jimmy Lei and Kiros, Jamie Ryan and Hinton, Geoffrey E.},
  title         = {Layer Normalization},
  journal       = {arXiv preprint arXiv:1607.06450},
  year          = {2016},
  eprint        = {1607.06450},
  archiveprefix = {arXiv},
  primaryclass  = {stat.ML},
  url           = {https://arxiv.org/abs/1607.06450}
}

@article{rumelhart1986learning,
  author  = {Rumelhart, David E. and Hinton, Geoffrey E. and Williams, Ronald J.},
  title   = {Learning Representations by Back-Propagating Errors},
  journal = {Nature},
  volume  = {323},
  number  = {6088},
  pages   = {533--536},
  year    = {1986},
  doi     = {10.1038/323533a0}
}

@article{rosenblatt1958perceptron,
  author  = {Rosenblatt, Frank},
  title   = {The Perceptron: A Probabilistic Model for Information Storage and Organization in the Brain},
  journal = {Psychological Review},
  volume  = {65},
  number  = {6},
  pages   = {386--408},
  year    = {1958},
  doi     = {10.1037/h0042519}
}

@article{fukushima1980neocognitron,
  author  = {Fukushima, Kunihiko},
  title   = {Neocognitron: A Self-Organizing Neural Network Model for a Mechanism of Pattern Recognition Unaffected by Shift in Position},
  journal = {Biological Cybernetics},
  volume  = {36},
  number  = {4},
  pages   = {193--202},
  year    = {1980},
  doi     = {10.1007/BF00344251}
}

@article{lecun1998gradient,
  author  = {LeCun, Yann and Bottou, L{\'e}on and Bengio, Yoshua and Haffner, Patrick},
  title   = {Gradient-Based Learning Applied to Document Recognition},
  journal = {Proceedings of the IEEE},
  volume  = {86},
  number  = {11},
  pages   = {2278--2324},
  year    = {1998},
  doi     = {10.1109/5.726791}
}

@article{elman1990finding,
  author  = {Elman, Jeffrey L.},
  title   = {Finding Structure in Time},
  journal = {Cognitive Science},
  volume  = {14},
  number  = {2},
  pages   = {179--211},
  year    = {1990},
  doi     = {10.1207/s15516709cog1402_1}
}

@inproceedings{cho2014learning,
  author    = {Cho, Kyunghyun and van Merri{\"e}nboer, Bart and Gulcehre, Caglar and Bahdanau, Dzmitry and Bougares, Fethi and Schwenk, Holger and Bengio, Yoshua},
  title     = {Learning Phrase Representations using {RNN} Encoder--Decoder for Statistical Machine Translation},
  booktitle = {Proceedings of the 2014 Conference on Empirical Methods in Natural Language Processing ({EMNLP})},
  year      = {2014},
}

@article{poli2024mechanistic,
  title={Mechanistic design and scaling of hybrid architectures},
  author={Poli, Michael and Thomas, Armin W and Nguyen, Eric and Ponnusamy, Pragaash and Deiseroth, Bj{\"o}rn and Kersting, Kristian and Suzuki, Taiji and Hie, Brian and Ermon, Stefano and R{\'e}, Christopher and others},
  journal={arXiv preprint arXiv:2403.17844},
  year={2024}
}

@article{hoffmann2022training,
  title={Training compute-optimal large language models},
  author={Hoffmann, Jordan and Borgeaud, Sebastian and Mensch, Arthur and Buchatskaya, Elena and Cai, Trevor and Rutherford, Eliza and Casas, DDL and Hendricks, Lisa Anne and Welbl, Johannes and Clark, Aidan and others},
  journal={arXiv preprint arXiv:2203.15556},
  volume={10},
  year={2022}
}

@article{hochreiter1997lstm,
  author  = {Hochreiter, Sepp and Schmidhuber, J{\"u}rgen},
  title   = {Long Short-Term Memory},
  journal = {Neural Computation},
  volume  = {9},
  number  = {8},
  pages   = {1735--1780},
  year    = {1997},
  doi     = {10.1162/neco.1997.9.8.1735}
}

@article{bahdanau2014nmtattention,
  author        = {Bahdanau, Dzmitry and Cho, Kyunghyun and Bengio, Yoshua},
  title         = {Neural Machine Translation by Jointly Learning to Align and Translate},
  journal       = {arXiv preprint arXiv:1409.0473},
  year          = {2014},
  eprint        = {1409.0473},
  archiveprefix = {arXiv},
  primaryclass  = {cs.CL},
  url           = {https://arxiv.org/abs/1409.0473}
}

@inproceedings{vaswani2017attention,
  author    = {Vaswani, Ashish and Shazeer, Noam and Parmar, Niki and Uszkoreit, Jakob and Jones, Llion and Gomez, Aidan N. and Kaiser, {\L}ukasz and Polosukhin, Illia},
  title     = {Attention Is All You Need},
  booktitle = {Advances in Neural Information Processing Systems 30},
  year      = {2017},
}

@inproceedings{gu2022efficiently,
  title={Efficiently modeling long sequences with structured state spaces},
  author={Gu, Albert and Goel, Karan and R{\'e}, Christopher},
booktitle={Proceedings of ICLR},
  year={2022}
}

@inproceedings{Gu2023MambaLS,
 author = {Albert Gu and Tri Dao},
 title = {Mamba: Linear-Time Sequence Modeling with Selective State Spaces},
 year = {2024},
  booktitle={Proceedings of CoLM},
}

@inproceedings{yang2024gated,
      title={Gated Delta Networks: Improving Mamba2 with Delta Rule}, 
      author={Songlin Yang and Jan Kautz and Ali Hatamizadeh},
 booktitle = {Proceedings of ICLR},
 year = {2025}
}

@inproceedings{schlag_linear_2021,
	title = {Linear {Transformers} {Are} {Secretly} {Fast} {Weight} {Programmers}},
  author={Schlag, Imanol and Irie, Kazuki and Schmidhuber, J{\"u}rgen},
	booktitle = {Proceedings of ICML},
	year = {2021},
}

@inproceedings{yang2024parallelizing,
  title     = {Parallelizing Linear Transformers with the Delta Rule over Sequence Length},
  author    = {Yang, Songlin and Wang, Bailin and Zhang, Yu and Shen, Yikang and Kim, Yoon},
  booktitle = {Proceedings of NeurIPS},
  year      = {2024}
}

@misc{yang2025qwen3technicalreport,
      title={Qwen3 Technical Report}, 
      author={An Yang and Anfeng Li and Baosong Yang and Beichen Zhang and Binyuan Hui and Bo Zheng and Bowen Yu and Chang Gao and Chengen Huang and Chenxu Lv and Chujie Zheng and Dayiheng Liu and Fan Zhou and Fei Huang and Feng Hu and Hao Ge and Haoran Wei and Huan Lin and Jialong Tang and Jian Yang and Jianhong Tu and Jianwei Zhang and Jianxin Yang and Jiaxi Yang and Jing Zhou and Jingren Zhou and Junyang Lin and Kai Dang and Keqin Bao and Kexin Yang and Le Yu and Lianghao Deng and Mei Li and Mingfeng Xue and Mingze Li and Pei Zhang and Peng Wang and Qin Zhu and Rui Men and Ruize Gao and Shixuan Liu and Shuang Luo and Tianhao Li and Tianyi Tang and Wenbiao Yin and Xingzhang Ren and Xinyu Wang and Xinyu Zhang and Xuancheng Ren and Yang Fan and Yang Su and Yichang Zhang and Yinger Zhang and Yu Wan and Yuqiong Liu and Zekun Wang and Zeyu Cui and Zhenru Zhang and Zhipeng Zhou and Zihan Qiu},
      year={2025},
      eprint={2505.09388},
      archivePrefix={arXiv},
      primaryClass={cs.CL},
      url={https://arxiv.org/abs/2505.09388}, 
}

@misc{5team2025glm45agenticreasoningcoding,
      title={GLM-4.5: Agentic, Reasoning, and Coding (ARC) Foundation Models}, 
      author={ 5 Team and Aohan Zeng and Xin Lv and Qinkai Zheng and Zhenyu Hou and Bin Chen and Chengxing Xie and Cunxiang Wang and Da Yin and Hao Zeng and Jiajie Zhang and Kedong Wang and Lucen Zhong and Mingdao Liu and Rui Lu and Shulin Cao and Xiaohan Zhang and Xuancheng Huang and Yao Wei and Yean Cheng and Yifan An and Yilin Niu and Yuanhao Wen and Yushi Bai and Zhengxiao Du and Zihan Wang and Zilin Zhu and Bohan Zhang and Bosi Wen and Bowen Wu and Bowen Xu and Can Huang and Casey Zhao and Changpeng Cai and Chao Yu and Chen Li and Chendi Ge and Chenghua Huang and Chenhui Zhang and Chenxi Xu and Chenzheng Zhu and Chuang Li and Congfeng Yin and Daoyan Lin and Dayong Yang and Dazhi Jiang and Ding Ai and Erle Zhu and Fei Wang and Gengzheng Pan and Guo Wang and Hailong Sun and Haitao Li and Haiyang Li and Haiyi Hu and Hanyu Zhang and Hao Peng and Hao Tai and Haoke Zhang and Haoran Wang and Haoyu Yang and He Liu and He Zhao and Hongwei Liu and Hongxi Yan and Huan Liu and Huilong Chen and Ji Li and Jiajing Zhao and Jiamin Ren and Jian Jiao and Jiani Zhao and Jianyang Yan and Jiaqi Wang and Jiayi Gui and Jiayue Zhao and Jie Liu and Jijie Li and Jing Li and Jing Lu and Jingsen Wang and Jingwei Yuan and Jingxuan Li and Jingzhao Du and Jinhua Du and Jinxin Liu and Junkai Zhi and Junli Gao and Ke Wang and Lekang Yang and Liang Xu and Lin Fan and Lindong Wu and Lintao Ding and Lu Wang and Man Zhang and Minghao Li and Minghuan Xu and Mingming Zhao and Mingshu Zhai and Pengfan Du and Qian Dong and Shangde Lei and Shangqing Tu and Shangtong Yang and Shaoyou Lu and Shijie Li and Shuang Li and Shuang-Li and Shuxun Yang and Sibo Yi and Tianshu Yu and Wei Tian and Weihan Wang and Wenbo Yu and Weng Lam Tam and Wenjie Liang and Wentao Liu and Xiao Wang and Xiaohan Jia and Xiaotao Gu and Xiaoying Ling and Xin Wang and Xing Fan and Xingru Pan and Xinyuan Zhang and Xinze Zhang and Xiuqing Fu and Xunkai Zhang and Yabo Xu and Yandong Wu and Yida Lu and Yidong Wang and Yilin Zhou and Yiming Pan and Ying Zhang and Yingli Wang and Yingru Li and Yinpei Su and Yipeng Geng and Yitong Zhu and Yongkun Yang and Yuhang Li and Yuhao Wu and Yujiang Li and Yunan Liu and Yunqing Wang and Yuntao Li and Yuxuan Zhang and Zezhen Liu and Zhen Yang and Zhengda Zhou and Zhongpei Qiao and Zhuoer Feng and Zhuorui Liu and Zichen Zhang and Zihan Wang and Zijun Yao and Zikang Wang and Ziqiang Liu and Ziwei Chai and Zixuan Li and Zuodong Zhao and Wenguang Chen and Jidong Zhai and Bin Xu and Minlie Huang and Hongning Wang and Juanzi Li and Yuxiao Dong and Jie Tang},
      year={2025},
      eprint={2508.06471},
      archivePrefix={arXiv},
      primaryClass={cs.CL},
      url={https://arxiv.org/abs/2508.06471}, 
}

@misc{kaplan2020scalinglawsneurallanguage,
      title={Scaling Laws for Neural Language Models}, 
      author={Jared Kaplan and Sam McCandlish and Tom Henighan and Tom B. Brown and Benjamin Chess and Rewon Child and Scott Gray and Alec Radford and Jeffrey Wu and Dario Amodei},
      year={2020},
      eprint={2001.08361},
      archivePrefix={arXiv},
      primaryClass={cs.LG},
      url={https://arxiv.org/abs/2001.08361}, 
}

@inproceedings{tillet2019triton,
  title={Triton: an intermediate language and compiler for tiled neural network computations},
  author={Tillet, Philippe and Kung, Hsiang-Tsung and Cox, David},
  booktitle={Proceedings of the 3rd ACM SIGPLAN International Workshop on Machine Learning and Programming Languages},
  pages={10--19},
  year={2019}
}

@InProceedings{pmlr-v202-dehghani23a,
  title = 	 {Scaling Vision Transformers to 22 Billion Parameters},
  author =       {Dehghani, Mostafa and Djolonga, Josip and Mustafa, Basil and Padlewski, Piotr and Heek, Jonathan and Gilmer, Justin and Steiner, Andreas Peter and Caron, Mathilde and Geirhos, Robert and Alabdulmohsin, Ibrahim and Jenatton, Rodolphe and Beyer, Lucas and Tschannen, Michael and Arnab, Anurag and Wang, Xiao and Riquelme Ruiz, Carlos and Minderer, Matthias and Puigcerver, Joan and Evci, Utku and Kumar, Manoj and Steenkiste, Sjoerd Van and Elsayed, Gamaleldin Fathy and Mahendran, Aravindh and Yu, Fisher and Oliver, Avital and Huot, Fantine and Bastings, Jasmijn and Collier, Mark and Gritsenko, Alexey A. and Birodkar, Vighnesh and Vasconcelos, Cristina Nader and Tay, Yi and Mensink, Thomas and Kolesnikov, Alexander and Pavetic, Filip and Tran, Dustin and Kipf, Thomas and Lucic, Mario and Zhai, Xiaohua and Keysers, Daniel and Harmsen, Jeremiah J. and Houlsby, Neil},
  booktitle = 	 {Proceedings of the 40th International Conference on Machine Learning},
  pages = 	 {7480--7512},
  year = 	 {2023},
  editor = 	 {Krause, Andreas and Brunskill, Emma and Cho, Kyunghyun and Engelhardt, Barbara and Sabato, Sivan and Scarlett, Jonathan},
  volume = 	 {202},
  series = 	 {Proceedings of Machine Learning Research},
  month = 	 {23--29 Jul},
  publisher =    {PMLR},
  url = 	 {https://proceedings.mlr.press/v202/dehghani23a.html}
}

@article{arora2023zoology,
  title={Zoology: Measuring and improving recall in efficient language models},
  author={Arora, Simran and Eyuboglu, Sabri and Timalsina, Aman and Johnson, Isys and Poli, Michael and Zou, James and Rudra, Atri and R{\'e}, Christopher},
  journal={arXiv preprint arXiv:2312.04927},
  year={2023}
}

@misc{gemmateam2025gemma3technicalreport,
      title={Gemma 3 Technical Report}, 
      author={Gemma Team and Aishwarya Kamath and Johan Ferret and Shreya Pathak and Nino Vieillard and Ramona Merhej and Sarah Perrin and Tatiana Matejovicova and Alexandre Ramé and Morgane Rivière and Louis Rouillard and Thomas Mesnard and Geoffrey Cideron and Jean-bastien Grill and Sabela Ramos and Edouard Yvinec and Michelle Casbon and Etienne Pot and Ivo Penchev and Gaël Liu and Francesco Visin and Kathleen Kenealy and Lucas Beyer and Xiaohai Zhai and Anton Tsitsulin and Robert Busa-Fekete and Alex Feng and Noveen Sachdeva and Benjamin Coleman and Yi Gao and Basil Mustafa and Iain Barr and Emilio Parisotto and David Tian and Matan Eyal and Colin Cherry and Jan-Thorsten Peter and Danila Sinopalnikov and Surya Bhupatiraju and Rishabh Agarwal and Mehran Kazemi and Dan Malkin and Ravin Kumar and David Vilar and Idan Brusilovsky and Jiaming Luo and Andreas Steiner and Abe Friesen and Abhanshu Sharma and Abheesht Sharma and Adi Mayrav Gilady and Adrian Goedeckemeyer and Alaa Saade and Alex Feng and Alexander Kolesnikov and Alexei Bendebury and Alvin Abdagic and Amit Vadi and András György and André Susano Pinto and Anil Das and Ankur Bapna and Antoine Miech and Antoine Yang and Antonia Paterson and Ashish Shenoy and Ayan Chakrabarti and Bilal Piot and Bo Wu and Bobak Shahriari and Bryce Petrini and Charlie Chen and Charline Le Lan and Christopher A. Choquette-Choo and CJ Carey and Cormac Brick and Daniel Deutsch and Danielle Eisenbud and Dee Cattle and Derek Cheng and Dimitris Paparas and Divyashree Shivakumar Sreepathihalli and Doug Reid and Dustin Tran and Dustin Zelle and Eric Noland and Erwin Huizenga and Eugene Kharitonov and Frederick Liu and Gagik Amirkhanyan and Glenn Cameron and Hadi Hashemi and Hanna Klimczak-Plucińska and Harman Singh and Harsh Mehta and Harshal Tushar Lehri and Hussein Hazimeh and Ian Ballantyne and Idan Szpektor and Ivan Nardini and Jean Pouget-Abadie and Jetha Chan and Joe Stanton and John Wieting and Jonathan Lai and Jordi Orbay and Joseph Fernandez and Josh Newlan and Ju-yeong Ji and Jyotinder Singh and Kat Black and Kathy Yu and Kevin Hui and Kiran Vodrahalli and Klaus Greff and Linhai Qiu and Marcella Valentine and Marina Coelho and Marvin Ritter and Matt Hoffman and Matthew Watson and Mayank Chaturvedi and Michael Moynihan and Min Ma and Nabila Babar and Natasha Noy and Nathan Byrd and Nick Roy and Nikola Momchev and Nilay Chauhan and Noveen Sachdeva and Oskar Bunyan and Pankil Botarda and Paul Caron and Paul Kishan Rubenstein and Phil Culliton and Philipp Schmid and Pier Giuseppe Sessa and Pingmei Xu and Piotr Stanczyk and Pouya Tafti and Rakesh Shivanna and Renjie Wu and Renke Pan and Reza Rokni and Rob Willoughby and Rohith Vallu and Ryan Mullins and Sammy Jerome and Sara Smoot and Sertan Girgin and Shariq Iqbal and Shashir Reddy and Shruti Sheth and Siim Põder and Sijal Bhatnagar and Sindhu Raghuram Panyam and Sivan Eiger and Susan Zhang and Tianqi Liu and Trevor Yacovone and Tyler Liechty and Uday Kalra and Utku Evci and Vedant Misra and Vincent Roseberry and Vlad Feinberg and Vlad Kolesnikov and Woohyun Han and Woosuk Kwon and Xi Chen and Yinlam Chow and Yuvein Zhu and Zichuan Wei and Zoltan Egyed and Victor Cotruta and Minh Giang and Phoebe Kirk and Anand Rao and Kat Black and Nabila Babar and Jessica Lo and Erica Moreira and Luiz Gustavo Martins and Omar Sanseviero and Lucas Gonzalez and Zach Gleicher and Tris Warkentin and Vahab Mirrokni and Evan Senter and Eli Collins and Joelle Barral and Zoubin Ghahramani and Raia Hadsell and Yossi Matias and D. Sculley and Slav Petrov and Noah Fiedel and Noam Shazeer and Oriol Vinyals and Jeff Dean and Demis Hassabis and Koray Kavukcuoglu and Clement Farabet and Elena Buchatskaya and Jean-Baptiste Alayrac and Rohan Anil and Dmitry and Lepikhin and Sebastian Borgeaud and Olivier Bachem and Armand Joulin and Alek Andreev and Cassidy Hardin and Robert Dadashi and Léonard Hussenot},
      year={2025},
      eprint={2503.19786},
      archivePrefix={arXiv},
      primaryClass={cs.CL},
      url={https://arxiv.org/abs/2503.19786}, 
}

@inproceedings{paperno_lambada_2016,
 address = {Berlin, Germany},
 author = {Paperno, Denis  and
Kruszewski, Germ{\'a}n  and
Lazaridou, Angeliki  and
Pham, Ngoc Quan  and
Bernardi, Raffaella  and
Pezzelle, Sandro  and
Baroni, Marco  and
Boleda, Gemma  and
Fern{\'a}ndez, Raquel},
 booktitle = {Proceedings of the 54th Annual Meeting of the Association for Computational Linguistics (Volume 1: Long Papers)},
 doi = {10.18653/v1/P16-1144},
 editor = {Erk, Katrin  and
Smith, Noah A.},
 pages = {1525--1534},
 publisher = {Association for Computational Linguistics},
 title = {The {LAMBADA} dataset: Word prediction requiring a broad discourse context},
 url = {https://aclanthology.org/P16-1144},
 year = {2016}
}

@misc{eval-harness,
  author       = {Gao, Leo and Tow, Jonathan and Abbasi, Baber and Biderman, Stella and Black, Sid and DiPofi, Anthony and Foster, Charles and Golding, Laurence and Hsu, Jeffrey and Le Noac'h, Alain and Li, Haonan and McDonell, Kyle and Muennighoff, Niklas and Ociepa, Chris and Phang, Jason and Reynolds, Laria and Schoelkopf, Hailey and Skowron, Aviya and Sutawika, Lintang and Tang, Eric and Thite, Anish and Wang, Ben and Wang, Kevin and Zou, Andy},
  title        = {The Language Model Evaluation Harness},
  month        = 07,
  year         = 2024,
  publisher    = {Zenodo},
  version      = {v0.4.3},
  doi          = {10.5281/zenodo.12608602},
  url          = {https://zenodo.org/records/12608602}
}

\bibliographystyle{icml2026}

\appendix

\vspace{-2mm}
\section{Kernel Benchmark Setup}
\label{app:kernel_baselines}
All measurements were performed on a single NVIDIA H100 SXM5 80GB HBM3 GPU, which has a theoretical peak HBM bandwidth of $3.35$~TB/s and peak BF16 matrix-multiply throughput of $989$~TFLOPs. Our benchmarking environment used PyTorch 2.12.0 nightly with CUDA 13.0, cuDNN 9.2.0, Triton 3.7.0, and \texttt{causal\_conv1d} 1.6.1.

For both the head-wise and the low-rank variant we implement five mathematically equivalent variants in PyTorch. Listing~\ref{lst:headwise} contains the five variants of the head-wise implementation, and Listing~\ref{lst:lowrank} contains the five variants of the low-rank implementation. Each variant is benchmarked in PyTorch eager mode and under all four \texttt{torch.compile} modes (\texttt{default}, \texttt{reduce-overhead}, \texttt{max-autotune}, \texttt{max-autotune-no-cudagraphs}). All \texttt{torch.compile} runs are benchmarked with  \texttt{fullgraph=True}, \texttt{dynamic=False}, and Dynamo's recompile limit increased to $64$. We report the fastest formulation-mode combination in the main figure. Table~\ref{tab:kernel_full_results} contains the winning variant for torch/torch.compile in each setting.

\begin{lstlisting}[language=Python,caption={Five mathematically equivalent PyTorch implementations of the head-wise dynamic convolution. Here, $H$ denotes the head size and $G=D/H$ the number of groups, consistent with the main-text notation.},label={lst:headwise}]
def hw_loop_pad(x, weight):
    B, T, D = x.shape
    _, _, G, W = weight.shape
    H = D // G
    x_h = x.view(B, T, G, H)
    out = torch.zeros_like(x_h)
    for w in range(W):
        x_shift = F.pad(x_h, (0, 0, 0, 0, w, 0))[:, :T]
        out = out + weight[:, :, :, w:w + 1] * x_shift
    return out.reshape(B, T, D)


def hw_unfold(x, weight):
    B, T, D = x.shape
    _, _, G, W = weight.shape
    H = D // G
    x_h = x.view(B, T, G, H)
    x_pad = F.pad(x_h, (0, 0, 0, 0, W - 1, 0))
    windows = x_pad.unfold(1, W, 1).flip(-1)
    return (windows * weight.unsqueeze(-2)).sum(-1).reshape(B, T, D)


def hw_einsum(x, weight):
    B, T, D = x.shape
    _, _, G, W = weight.shape
    H = D // G
    x_h = x.view(B, T, G, H)
    x_pad = F.pad(x_h, (0, 0, 0, 0, W - 1, 0))
    windows = x_pad.unfold(1, W, 1).flip(-1)
    return torch.einsum('btghw,btgw->btgh', windows, weight).reshape(B, T, D)


def hw_stack(x, weight):
    B, T, D = x.shape
    _, _, G, W = weight.shape
    H = D // G
    x_h = x.view(B, T, G, H)
    shifts = [x_h]
    for w in range(1, W):
        zero = x_h.new_zeros(B, w, G, H)
        shifts.append(torch.cat([zero, x_h[:, :T - w]], dim=1))
    stacked = torch.stack(shifts, dim=-1)
    return (stacked * weight.unsqueeze(-2)).sum(-1).reshape(B, T, D)


def hw_bmm(x, weight):
    B, T, D = x.shape
    _, _, G, W = weight.shape
    H = D // G
    x_h = x.view(B, T, G, H)
    x_pad = F.pad(x_h, (0, 0, 0, 0, W - 1, 0))
    windows = x_pad.unfold(1, W, 1).flip(-1)
    BT_G = B * T * G
    out = torch.bmm(
        windows.reshape(BT_G, H, W),
        weight.reshape(BT_G, W, 1),
    )
    return out.reshape(B, T, D)
\end{lstlisting}

\newpage
\begin{lstlisting}[language=Python,caption={Five mathematically equivalent PyTorch implementations of the low-rank dynamic convolution. Latency includes the second projection of the low-rank factorization.},label={lst:lowrank}]
def lr_materialize_loop(x, z, U):
    B, T, D = x.shape
    R, WD = U.shape
    W = WD // D
    weight = (z @ U).view(B, T, W, D)
    out = torch.zeros_like(x)
    for w in range(W):
        x_shift = F.pad(x, (0, 0, w, 0))[:, :T]
        out = out + weight[:, :, w, :] * x_shift
    return out


def lr_materialize_unfold(x, z, U):
    B, T, D = x.shape
    R, WD = U.shape
    W = WD // D
    Ur = U.view(R, W, D)
    weight = torch.einsum('btr,rwd->btwd', z, Ur)
    x_pad = F.pad(x, (0, 0, W - 1, 0))
    windows = x_pad.unfold(1, W, 1).flip(-1)
    return (windows * weight.permute(0, 1, 3, 2)).sum(-1)


def lr_unfold_einsum(x, z, U):
    B, T, D = x.shape
    R, WD = U.shape
    W = WD // D
    Ur = U.view(R, W, D)
    x_pad = F.pad(x, (0, 0, W - 1, 0))
    windows = x_pad.unfold(1, W, 1).flip(-1)
    return torch.einsum('btdw,btr,rwd->btd', windows, z, Ur)


def lr_fused_per_tap(x, z, U):
    B, T, D = x.shape
    R, WD = U.shape
    W = WD // D
    Uw = U.view(R, W, D)
    out = torch.zeros_like(x)
    for w in range(W):
        weight_w = z @ Uw[:, w, :]
        x_shift = F.pad(x, (0, 0, w, 0))[:, :T]
        out = out + weight_w * x_shift
    return out


def lr_static_conv_per_rank(x, z, U):
    B, T, D = x.shape
    R, WD = U.shape
    W = WD // D
    Ur = U.view(R, W, D)
    Uf = Ur.flip(1)
    x_bdt = x.transpose(1, 2).contiguous()
    x_pad = F.pad(x_bdt, (W - 1, 0))
    out = torch.zeros(B, T, D, dtype=x.dtype, device=x.device)
    for r in range(R):
        weight_r = Uf[r].T.unsqueeze(1).contiguous()
        conv_r = F.conv1d(x_pad, weight_r, groups=D)
        out = out + z[:, :, r:r + 1] * conv_r.transpose(1, 2)
    return out
\end{lstlisting}

We benchmark using Triton's \texttt{triton.testing.do\_bench} with 500ms warmup and 3000ms measurement window, and record the median. We repeat this 5 times and report the run with the lowest median. We time the forward and the forward+backward independently and report the difference of the medians as the backward latency.

\begin{table}[h]
\centering
\setlength{\tabcolsep}{4pt}
\caption{Per-configuration latency at
$B{=}4$, $T{=}4096$, $D{=}2048$, $W{=}4$, BF16. All kernels were tested on $B{\times}T{\times}D$ activations with the $D$ dimension contiguous in memory. \texttt{torch.compile} cells are labeled \texttt{(<variant>, <Inductor mode>)}. The last row implements a static
convolution ($W{=}4$).}
\label{tab:kernel_full_results}
\resizebox{\linewidth}{!}{%
\begin{tabular}{l|l|ccc}
\toprule
Configuration & Implementation & fwd (ms) & bwd (ms) & fwd+bwd (ms) \\
\midrule
\multirow{3}{*}{head-wise: $H{=}1$}
  & triton                                                            & $\mathbf{0.140}$ & $\mathbf{0.243}$ & $\mathbf{0.382}$ \\
  & torch (unfold)                                                    & $1.022$ & $1.884$ & $2.906$ \\
  & torch.compile (stack, max-autotune-no-cudagraphs)                 & $0.330$ & $0.367$ & $0.697$ \\
\midrule
\multirow{3}{*}{head-wise: $H{=}4$}
  & triton                                                            & $\mathbf{0.073}$ & $\mathbf{0.111}$ & $\mathbf{0.184}$ \\
  & torch (unfold)                                                    & $1.033$ & $2.484$ & $3.517$ \\
  & torch.compile (stack, max-autotune-no-cudagraphs)                 & $0.272$ & $0.211$ & $0.484$ \\
\midrule
\multirow{3}{*}{head-wise: $H{=}16$}
  & triton                                                            & $\mathbf{0.055}$ & $\mathbf{0.088}$ & $\mathbf{0.143}$ \\
  & torch (unfold)                                                    & $1.029$ & $2.516$ & $3.545$ \\
  & torch.compile (loop\_pad, max-autotune-no-cudagraphs)             & $0.266$ & $0.155$ & $0.421$ \\
\midrule
\multirow{3}{*}{low-rank: $R{=}16$}
  & triton                                                            & $\mathbf{0.071}$ & $\mathbf{0.171}$ & $\mathbf{0.242}$ \\
  & torch (materialize\_unfold)                                       & $0.982$ & $1.729$ & $2.711$ \\
  & torch.compile (materialize\_loop, max-autotune-no-cudagraphs)     & $0.435$ & $0.511$ & $0.946$ \\
\midrule
static (cuda)
  & causal\_conv1d                                                    & $\mathbf{0.056}$ & $\mathbf{0.105}$ & $\mathbf{0.161}$ \\
\bottomrule
\end{tabular}%
}
\end{table}

\section{Full Evaluation Results}
\label{full_eval}

This section contains the per-task breakdowns for the downstream evaluations
summarized in Table~\ref{tab:main_eval_compact} of the main paper. The
zero-shot common-sense reasoning results are given in Table~\ref{tab:full_0shot}. For ARC-C/E,
HellaSwag, OpenBookQA, PIQA, and SciQ we report \texttt{acc\_norm}, and
\texttt{acc} for the remaining tasks. Results for all 11 non-QA subtasks of RULER (single/multi-key needle-in-a-haystack, multi-query/value NIAH, common-words extraction, frequent-words extraction, variable tracking) at context length $4096$
are provided in Table~\ref{tab:full_ruler_4k}.

 \begin{table}[thb]
\centering
\scriptsize
\setlength{\tabcolsep}{2.5pt}
\renewcommand{\arraystretch}{1.05}
\caption{Per-task 0-shot accuracy on the lm-eval-harness suite. For ARC-C/E, HellaSwag, OBQA, PIQA, SciQ we report \texttt{acc\_norm}, and for BoolQ, COPA, LAMBADA, RACE, WinoGrande we report \texttt{acc}. \emph{Avg.} is the mean of all tasks. }
\label{tab:full_0shot}
\resizebox{\linewidth}{!}{
\begin{tabular}{l|c|ccccccccccc|c}
\toprule
\textbf{Model} & \textbf{Params} & \textbf{ARC-C} & \textbf{ARC-E} & \textbf{BoolQ} & \textbf{COPA} & \textbf{Hella.} & \textbf{LAMB.} & \textbf{OBQA} & \textbf{PIQA} & \textbf{RACE} & \textbf{SciQ} & \textbf{WinoG.} & \textbf{Avg.} \\
\midrule

MoE Transformer (\emph{100B Tokens})
& 6.77B
& 44.54 & 72.73 & 65.29 & 80.00 & 68.08 & 49.66 & 41.00 & 77.69 & \textbf{36.75} & 90.80 & 60.54 & 62.46 \\

w/ static conv.
& 6.77B
& 44.20 & 73.40 & 65.66 & \textbf{82.00} & 68.60 & 50.75 & 41.60 & 77.42 & 36.27 & 91.90 & 60.85 & 62.97 \\

w/ dynamic conv. (head-wise)
& 6.80B
& \textbf{45.14} & \textbf{74.20} & 67.46 & 80.00 & 69.12 & \textbf{51.31} & \textbf{43.20} & 78.02 & 35.02 & 91.10 & \textbf{63.14} & \textbf{63.43} \\

w/ dynamic conv. (low-rank)
& 6.80B
& 44.88 & 74.03 & \textbf{68.13} & 81.00 & \textbf{69.22} & 51.27 & 42.80 & \textbf{78.51} & 35.69 & \textbf{92.20} & 59.91 & 63.42 \\

\midrule

Transformer (\emph{100B Tokens})
& 1.82B
& 38.14 & 68.31 & 60.49 & 78.00 & 60.49 & 44.36 & 38.60 & 74.86 & 34.83 & 87.00 & 56.75 & 58.35 \\

w/ more params
& 1.87B
& 37.54 & 67.85 & \textbf{63.67} & 77.00 & 60.40 & 43.51 & 38.40 & 74.76 & 34.74 & 86.20 & 56.51 & 58.23 \\

w/ static conv.
& 1.83B
& 37.97 & 68.27 & 61.96 & \textbf{79.00} & 61.69 & 45.06 & 38.60 & 74.65 & 34.55 & 89.50 & 57.14 & 58.94 \\

w/ dynamic conv. (head-wise)
& 1.88B
& 38.82 & 68.69 & 59.14 & 75.00 & 61.80 & 45.55 & 37.80 & 75.35 & 34.64 & 88.70 & 57.62 & 58.46 \\

w/ dynamic conv. (low-rank)
& 1.88B
& 39.16 & \textbf{69.49} & 63.64 & \textbf{79.00} & 62.57 & 46.09 & 38.40 & 75.46 & \textbf{35.79} & 88.70 & 58.41 & 59.70 \\

w/ dynamic conv. (all linear)
& 1.88B
& \textbf{40.44} & 69.36 & 63.18 & 78.00 & \textbf{64.73} & \textbf{48.67} & \textbf{41.00} & \textbf{76.01} & 34.83 & \textbf{91.00} & \textbf{60.46} & \textbf{60.70} \\

\midrule

Transformer (\emph{15B Tokens})
& 305.2M
& 26.79 & 52.61 & \textbf{60.52} & 63.00 & 37.40 & 26.96 & 30.80 & 66.27 & 30.33 & 73.60 & 51.62 & 47.26 \\

w/ more params
& 311.5M
& 26.45 & 51.47 & 52.23 & 63.00 & 37.56 & 29.17 & 32.00 & \textbf{68.34} & 29.47 & 73.10 & 50.20 & 46.64 \\

w/ static conv.
& 305.4M
& 27.22 & 51.18 & 53.79 & 62.00 & 38.35 & 28.66 & 31.20 & 66.76 & 30.33 & 75.60 & 50.36 & 46.86 \\

w/ dynamic conv. (head-wise)
& 311.7M
& 27.05 & 51.01 & 54.86 & 67.00 & 38.93 & 30.12 & 31.80 & 66.70 & \textbf{31.20} & 74.30 & 50.75 & 47.61 \\

w/ dynamic conv. (low-rank)
& 311.8M
& \textbf{27.90} & 52.57 & 58.01 & \textbf{73.00} & 39.90 & 29.77 & 32.00 & 67.03 & 30.81 & 76.00 & 50.91 & \textbf{48.90} \\

w/ dynamic conv. (all linear)
& 319.0M
& 27.56 & \textbf{54.46} & 56.18 & 68.00 & \textbf{41.02} & \textbf{30.51} & \textbf{32.20} & 67.68 & 31.10 & \textbf{76.10} & \textbf{52.09} & 48.81 \\

\midrule

Gated DeltaNet (w/o conv.)
& 305.2M
& 26.96 & 51.35 & 50.46 & 69.00 & 38.92 & 26.26 & \textbf{31.60} & 66.70 & 29.38 & 72.50 & \textbf{53.12} & 46.93 \\

w/ static conv.
& 305.4M
& 27.22 & 50.76 & 53.24 & 66.00 & 38.93 & 26.57 & 29.80 & 66.43 & 28.23 & 75.70 & 51.46 & 46.76 \\

w/ dynamic conv. (head-wise)
& 309.6M
& 26.71 & 53.37 & 47.89 & \textbf{70.00} & 40.27 & 26.84 & 31.20 & 67.36 & 29.00 & 75.50 & 51.78 & 47.27 \\

w/ dynamic conv. (low-rank)
& 309.5M
& \textbf{27.90} & \textbf{53.49} & \textbf{58.96} & \textbf{70.00} & \textbf{40.67} & \textbf{30.29} & 31.40 & \textbf{67.74} & \textbf{31.10} & \textbf{77.30} & 52.57 & \textbf{49.22} \\

\midrule

Mamba-2 (w/o conv.)
& 306.2M
& 26.96 & 49.49 & 57.80 & 70.00 & 36.72 & 24.08 & 30.60 & 66.65 & 29.19 & 70.10 & 48.93 & 46.41 \\

w/ static conv.
& 306.4M
& 25.94 & \textbf{52.19} & 48.84 & 65.00 & 38.67 & 23.97 & 31.40 & 66.87 & 28.33 & 72.50 & 49.88 & 45.78 \\

w/ dynamic conv. (head-wise)
& 309.8M
& \textbf{27.47} & 50.29 & \textbf{58.56} & 66.00 & 39.09 & 26.26 & 31.60 & \textbf{67.74} & 27.85 & \textbf{74.80} & \textbf{49.96} & 47.24 \\

w/ dynamic conv. (low-rank)
& 309.8M
& 25.94 & 51.43 & 54.71 & \textbf{71.00} & \textbf{39.10} & \textbf{26.51} & \textbf{32.40} & 65.89 & \textbf{29.76} & 74.40 & 49.57 & \textbf{47.34} \\

\bottomrule
\end{tabular}
}
\end{table}


\begin{table}[htb]

\centering
\scriptsize
\setlength{\tabcolsep}{2.5pt}
\renewcommand{\arraystretch}{1.05}
\caption{Per-subtask RULER accuracy at context length 4096.}
\label{tab:full_ruler_4k}
\resizebox{\linewidth}{!}{
\begin{tabular}{l|c|ccccccccccc|c}
\toprule
\textbf{Model} & \textbf{Params} & \textbf{S1} & \textbf{S2} & \textbf{S3} & \textbf{MK1} & \textbf{MK2} & \textbf{MK3} & \textbf{MQ} & \textbf{MV} & \textbf{CWE} & \textbf{FWE} & \textbf{VT} & \textbf{Avg.} \\
\midrule

MoE Transformer (\emph{100B Tokens})
& 6.77B
& 99.8 & \textbf{100.0} & 83.8 & 70.0 & 4.8 & 21.2 & 39.4 & 32.0 & 26.4 & \textbf{43.3} & 9.3 & 48.2 \\

w/ static conv.
& 6.77B
& \textbf{100.0} & \textbf{100.0} & 73.8 & 72.0 & 27.6 & 6.0 & 36.5 & 40.8 & 13.3 & 26.8 & \textbf{23.9} & 47.3 \\

w/ dynamic conv. (head-wise)
& 6.80B
& \textbf{100.0} & 99.8 & 81.8 & \textbf{74.2} & \textbf{45.4} & \textbf{38.4} & 37.4 & 40.5 & 15.1 & 16.3 & 8.1 & 50.6 \\

w/ dynamic conv. (low-rank)
& 6.80B
& 99.8 & \textbf{100.0} & \textbf{93.0} & 66.2 & 12.8 & 11.8 & \textbf{48.2} & \textbf{50.4} & \textbf{31.1} & 26.4 & 18.4 & \textbf{50.7} \\

\midrule

Transformer (\emph{100B Tokens})
& 1.82B
& \textbf{100.0} & \textbf{100.0} & 80.8 & 57.8 & 1.2 & 3.8 & 39.6 & 37.4 & 10.5 & 32.1 & 4.0 & 42.5 \\

w/ more params
& 1.87B
& \textbf{100.0} & 97.8 & 71.6 & 59.8 & 0.8 & 2.0 & 30.6 & 35.0 & 21.5 & 25.9 & 3.3 & 40.8 \\

w/ static conv.
& 1.83B
& 99.4 & 94.8 & 79.6 & 68.0 & 3.0 & 4.0 & \textbf{48.2} & \textbf{49.5} & 6.1 & \textbf{37.9} & \textbf{20.0} & \textbf{46.4} \\

w/ dynamic conv. (head-wise)
& 1.88B
& \textbf{100.0} & \textbf{100.0} & \textbf{95.6} & 61.0 & \textbf{7.4} & 6.0 & 24.9 & 28.6 & 13.3 & 36.3 & 8.7 & 43.8 \\

w/ dynamic conv. (low-rank)
& 1.88B
& \textbf{100.0} & 99.8 & 80.6 & 52.6 & 3.0 & 3.2 & 18.0 & 10.4 & 25.3 & 31.9 & 16.2 & 40.1 \\

w/ dynamic conv. (all linear)
& 1.88B
& \textbf{100.0} & \textbf{100.0} & 70.8 & \textbf{77.6} & 6.0 & \textbf{8.4} & 33.8 & 34.9 & \textbf{30.4} & \textbf{37.9} & 6.9 & 46.1 \\

\midrule

Transformer (\emph{15B Tokens})
& 305.2M
& 67.2 & 54.2 & 57.0 & 31.4 & 0.4 & 0.0 & 18.5 & 16.4 & 3.3 & 0.2 & 0.0 & 22.6 \\

w/ more params
& 311.5M
& 80.2 & 66.0 & 41.2 & 32.0 & 0.0 & 0.0 & 20.0 & 19.8 & 5.6 & 1.2 & 0.0 & 24.2 \\

w/ static conv.
& 305.4M
& 78.8 & 41.0 & 45.2 & 21.2 & 0.0 & 0.4 & 12.0 & 12.3 & 5.7 & \textbf{2.3} & 0.0 & 19.9 \\

w/ dynamic conv. (head-wise)
& 311.7M
& 97.0 & 69.4 & 68.0 & 31.6 & 0.2 & \textbf{1.0} & \textbf{22.7} & \textbf{21.3} & 0.0 & 1.4 & 0.5 & 28.5 \\

w/ dynamic conv. (low-rank)
& 311.8M
& 68.4 & 76.8 & \textbf{74.4} & 33.8 & \textbf{1.0} & \textbf{1.0} & 16.9 & 14.2 & \textbf{11.6} & 2.2 & 0.0 & 27.3 \\

w/ dynamic conv. (all linear)
& 319.0M
& \textbf{100.0} & \textbf{84.8} & 73.0 & \textbf{48.4} & 0.2 & 0.2 & 10.8 & 12.4 & 1.7 & 0.0 & \textbf{1.7} & \textbf{30.3} \\

\midrule

Gated DeltaNet (w/o conv.)
& 305.2M
& 90.6 & 35.8 & 22.2 & \textbf{19.6} & 0.0 & \textbf{0.0} & \textbf{16.4} & 11.1 & 2.5 & 0.7 & 0.1 & \textbf{18.1} \\

w/ static conv.
& 305.4M
& 99.2 & 33.0 & 4.8 & 19.2 & 0.0 & \textbf{0.0} & 2.8 & 2.8 & 0.9 & \textbf{5.5} & \textbf{5.0} & 15.7 \\

w/ dynamic conv. (head-wise)
& 309.6M
& \textbf{100.0} & 34.2 & \textbf{31.4} & 11.6 & 0.0 & \textbf{0.0} & 12.2 & 4.0 & 0.3 & 1.9 & 3.8 & \textbf{18.1} \\

w/ dynamic conv. (low-rank)
& 309.5M
& 99.8 & \textbf{36.0} & 7.2 & 18.4 & \textbf{0.2} & \textbf{0.0} & 9.7 & \textbf{16.4} & \textbf{5.3} & 2.9 & 2.1 & 18.0 \\

\midrule

Mamba-2 (w/o conv.)
& 306.2M
& 7.6 & 1.8 & 2.8 & 12.6 & \textbf{0.0} & \textbf{0.0} & 7.0 & 7.2 & \textbf{2.5} & 1.3 & \textbf{0.0} & 3.9 \\

w/ static conv.
& 306.4M
& 15.6 & 2.4 & 0.4 & 6.2 & \textbf{0.0} & \textbf{0.0} & 2.3 & 0.7 & 0.9 & 0.0 & \textbf{0.0} & 2.6 \\

w/ dynamic conv. (head-wise)
& 309.8M
& 30.2 & \textbf{17.4} & 18.4 & \textbf{21.2} & \textbf{0.0} & \textbf{0.0} & 11.9 & 9.2 & 0.6 & 2.2 & \textbf{0.0} & 10.1 \\

w/ dynamic conv. (low-rank)
& 309.8M
& \textbf{50.8} & 14.8 & \textbf{35.2} & 21.0 & \textbf{0.0} & \textbf{0.0} & \textbf{17.8} & \textbf{12.8} & 0.4 & \textbf{3.0} & \textbf{0.0} & \textbf{14.2} \\

\bottomrule
\end{tabular}
}
\end{table}


\section{Detailed Experimental Setup}
\label{app:detailed_hyperparams}

All models are trained in the \texttt{lm-engine} codebase \citep{mishra2024lmengine} on the Nemotron-CC corpus \citep{su-etal-2025-nemotron} tokenized with the Granite-4 BPE tokenizer (vocabulary $100{,}352$). All runs use sequence length $L=4096$, RMSNorm \citep{zhang2019rmsnorm} with $\varepsilon = 10^{-5}$, SwiGLU MLPs \citep{shazeer2020glu}, RoPE \citep{su2021roformer} on the full head dimension, untied input/output embeddings, no biases, and no dropout.
For optimization we use AdamW \citep{loshchilov2019decoupledweightdecayregularization} with $(\beta_1,\beta_2) = (0.9, 0.95)$, $\varepsilon = 10^{-10}$, weight decay $0.1$, and peak learning rate $3\!\times\!10^{-4}$ with $10\%$ warm-up and cosine decay to zero. Training uses bf16 mixed precision and \texttt{torch.compile} \citep{ansel2024pytorch2}.

Weights are initialized from $\mathcal{N}(0, 0.02^2)$ for all linear layers. Static convolution weights are initialized  to $\mathcal{U}(-1/\sqrt{W}, \; 1/\sqrt{W})$ per element (i.e., the default for \texttt{nn.Conv1d}). For the low-rank variant of dynamic convolutions we zero-initialize the second projection of the low-rank factorization and add a bias term to this projection, initialized  to $\mathcal{U}(-1/\sqrt{W}, \; 1/\sqrt{W})$ per element. We match this for the head-wise variant by zero-initializing the dynamic projection and adding a per-channel bias with the same Kaiming-uniform initialization. Through this, our dynamic convolutions match a static depthwise convolution at initialization.

Per-scale model architecture, batch size, and hardware are listed in Table~\ref{tab:hparams}. Dense model experiments use a token-to-parameter ratio of $\sim$$50$, and the 7B MoE uses 128 experts with top-$8$ routing, $256$ expert intermediate, and $1024$ shared-MLP intermediate. For $QKV$ placement the low-rank dynamic-convolution ranks $R$ are chosen so that the parameter count of the low-rank variant roughly matches the head-wise variant at head dimension $H=32$. For all-linear placement we set the rank to $R=16$. Convolution kernel width is $W=4$ throughout.

\begin{table}[t]

\centering
\scriptsize
\setlength{\tabcolsep}{3.0pt}
\renewcommand{\arraystretch}{1.08}
\caption{Architecture and training hyperparameters across model scales.  ``mbs'' is micro-batch size per device, ``gas'' is gradient-accumulation steps. All runs use NVIDIA H100 80GB HBM3 GPUs.}
\label{tab:hparams}
\resizebox{\linewidth}{!}{
\begin{tabular}{l|cccc|cccccc|cc}
\toprule
\multirow{2}{*}{\textbf{Scale}}
& \multicolumn{4}{c|}{\textbf{Architecture}}
& \multicolumn{6}{c|}{\textbf{Optimization}}
& \multicolumn{2}{c}{\textbf{Hardware}} \\
& \textbf{Layers}
& \textbf{$d_{\text{model}}$}
& \textbf{Heads}
& \textbf{MLP int.}
& \textbf{mbs}
& \textbf{gas}
& \textbf{Tot.\ bs.}
& \textbf{Eff.\ bs. (tok)}
& \textbf{Steps}
& \textbf{Tokens}
& \textbf{GPUs}
& \textbf{Nodes} \\
\midrule
150M & 12 &  768 & 12 & 2048 & 8 & 4 &  256 & $1.05$M &  8{,}000 &   8B & 8  & 1 \\
300M & 16 & 1024 & 16 & 2752 & 8 & 4 &  256 & $1.05$M & 15{,}000 &  15B & 8  & 1 \\
600M & 20 & 1280 & 20 & 3456 & 8 & 2 &  512 & $2.10$M & 13{,}000 &  27B & 32 & 4 \\
1B   & 26 & 1664 & 26 & 4480 & 4 & 4 &  512 & $2.10$M & 25{,}000 &  52B & 32 & 4 \\
2B   & 32 & 2048 & 32 & 5504 & 2 & 8 & 1024 & $4.19$M & 24{,}000 & 100B & 64 & 8 \\
\midrule
7B (MoE) & 40 & 1536 & 24 & 256 / 1024 & 2 & 8 & 1024 & $4.19$M & 25{,}000 & 100B & 64 & 8 \\
\bottomrule
\end{tabular}
}
\end{table}

\newpage
\section{Training Compute Convention}
\label{app:flop_calculations}

We report training compute as
\begin{equation}
    C_\text{full}
    \;=\;
    \underbrace{6 \cdot N_\text{ne} \cdot D}_{\text{matmul}}
    \;+\;
    \underbrace{12 \cdot n_\text{layers} \cdot d_\text{model} \cdot L \cdot D}_{\text{parameter-free attention}},
\end{equation}
where $N_\text{ne}$ is the non-embedding parameter count, $D$ is the number of tokens, $n_\text{layers}$ is the number of layers,  $d_\text{model}$ is the hidden dimension of the model, and $L$ is the sequence length. Since we are training on sequences of length $L=4096$, the parameter-free $QK^\top$ increases the total FLOPs significantly beyond the standard $6N$ matmul term \citep{kaplan2020scalinglawsneurallanguage, hoffmann2022training}. Across our sweep the attention term contributes 25-47\% of $C_\text{full}$.  We then convert to PFLOP-days via $1\,\text{PFLOP-day} = 8.64{\times}10^{19}\,\text{FLOPs}$. We report parameter counts and FLOPs of each model in Table~\ref{tab:flops}. The dynamic-convolution variants add $\sim 3\%$ to $N_\text{ne}$ through the weight-generation projections (low-rank factorizations). Notably, our advantage is robust to the training compute convention. Refitting under $C_\text{simple} = 6 N_\text{total} D$ or $C_\text{chinchilla} = 6 N_\text{ne} D$ gives compute advantages of $1.30\times$ and $1.34\times$ respectively (vs. $1.33\times$ for $C_\text{full}$).

\begin{table}[hbt]
\centering
\scriptsize
\setlength{\tabcolsep}{3.0pt}
\renewcommand{\arraystretch}{1.08}
\caption{Parameter counts and training compute for the dense scaling-law sweep.}
\label{tab:flops}
\resizebox{\linewidth}{!}{
\begin{tabular}{l|cc|ccc}
\toprule
\multirow{2}{*}{\textbf{Model}}
& \multicolumn{2}{c|}{\textbf{Parameters}}
& \multicolumn{3}{c}{\textbf{Compute}} \\
& \textbf{Total}
& \textbf{Non-emb.}
& \textbf{matmul (FLOPs)}
& \textbf{attention (FLOPs)}
& \textbf{$C_{\text{full}}$ (PFLOP-d)} \\
\midrule

Transformer (\emph{8B Tokens})
& 162.0M & 85.0M
& $4.28{\times}10^{18}$ & $3.80{\times}10^{18}$ & 0.09 \\

w/ dynamic conv. (low-rank)
& 164.9M & 87.8M
& $4.42{\times}10^{18}$ & $3.80{\times}10^{18}$ & 0.10 \\

w/ dynamic conv. (all linear)
& 169.8M & 92.7M
& $4.67{\times}10^{18}$ & $3.80{\times}10^{18}$ & 0.10 \\

\midrule

Transformer (\emph{15B Tokens})
& 305.2M & 202.4M
& $1.91{\times}10^{19}$ & $1.27{\times}10^{19}$ & 0.37 \\

w/ dynamic conv. (low-rank)
& 311.8M & 209.0M
& $1.97{\times}10^{19}$ & $1.27{\times}10^{19}$ & 0.37 \\

w/ dynamic conv. (all linear)
& 319.0M & 216.2M
& $2.04{\times}10^{19}$ & $1.27{\times}10^{19}$ & 0.38 \\

\midrule

Transformer (\emph{27B Tokens})
& 525.0M & 396.5M
& $6.49{\times}10^{19}$ & $3.43{\times}10^{19}$ & 1.15 \\

w/ dynamic conv. (low-rank)
& 537.6M & 409.1M
& $6.69{\times}10^{19}$ & $3.43{\times}10^{19}$ & 1.17 \\

w/ dynamic conv. (all linear)
& 546.7M & 418.2M
& $6.84{\times}10^{19}$ & $3.43{\times}10^{19}$ & 1.19 \\

\midrule

Transformer (\emph{52B Tokens})
& 1.04B & 869.5M
& $2.74{\times}10^{20}$ & $1.11{\times}10^{20}$ & 4.46 \\

w/ dynamic conv. (low-rank)
& 1.06B & 897.3M
& $2.82{\times}10^{20}$ & $1.11{\times}10^{20}$ & 4.56 \\

w/ dynamic conv. (all linear)
& 1.07B & 906.1M
& $2.85{\times}10^{20}$ & $1.11{\times}10^{20}$ & 4.59 \\

\midrule

Transformer (\emph{100B Tokens})
& 1.82B & 1.62B
& $9.78{\times}10^{20}$ & $3.24{\times}10^{20}$ & 15.07 \\

w/ dynamic conv. (low-rank)
& 1.88B & 1.67B
& $1.01{\times}10^{21}$ & $3.24{\times}10^{20}$ & 15.43 \\

w/ dynamic conv. (all linear)
& 1.88B & 1.67B
& $1.01{\times}10^{21}$ & $3.24{\times}10^{20}$ & 15.46 \\

\bottomrule
\end{tabular}
}
\end{table}



\end{document}